%% file: main.tex
\documentclass[10pt,twocolumn,letterpaper]{article}

\usepackage{cvpr}
\usepackage{times}
\usepackage{epsfig}
\usepackage{graphicx}
\usepackage{amsmath}
\usepackage{amssymb}
\usepackage{array}
\usepackage{float}
\usepackage[dvipsnames]{xcolor}
\newcommand*{\rowstyle}[1]{
	\gdef\@rowstyle{#1}
	\@rowstyle\ignorespaces%
}
\newcolumntype{=}{
	>{\gdef\@rowstyle{}}
}
\newcolumntype{+}{
	>{\@rowstyle}
}
\usepackage{caption}
\usepackage{subfigure}
\usepackage{xr}
\usepackage{url}

\usepackage[pagebackref=true,breaklinks=true,letterpaper=true,colorlinks,bookmarks=false]{hyperref}

\input{math_definition.tex}

\cvprfinalcopy 
\def\cvprPaperID{5326} 

\ifcvprfinal\fi
\begin{document}
	\title{End-to-End Pseudo-LiDAR for Image-Based 3D Object Detection}

	\author{Rui~Qian$^{*1,2}$\hspace{9pt} Divyansh~Garg$^{*1}$\hspace{9pt} Yan~Wang$^{*1}$\hspace{9pt} Yurong~You\thanks{\hspace{1pt}Equal contributions}\hspace{4pt}$^1$\\
		Serge~Belongie$^{1,2}$\hspace{9pt} Bharath~Hariharan$^{1}$\hspace{9pt} Mark~Campbell$^{1}$\hspace{9pt} Kilian~Q.~Weinberger$^{1}$\hspace{9pt} Wei-Lun~Chao$^{3}$ \\
		$^1$ Cornell Univeristy\hspace{9pt}$^2$ Cornell Tech\hspace{9pt}$^3$ The Ohio State University \\
		{\tt\small \{rq49, dg595, yw763, yy785, sjb344, bh497, mc288, kqw4\}@cornell.edu\hspace{5pt} chao.209@osu.edu}
	}

	\maketitle



	\input{abs.tex}
	\input{intro.tex}
	\input{related.tex}
	\input{approach.tex}
	\input{exp.tex}

	\input{disc.tex}
	\section*{Acknowledgments}
	{\small This research is supported by grants from the National Science Foundation NSF (III-1618134, III-1526012, IIS-1149882, IIS-1724282, and TRIPODS-1740822), the Office of Naval Research DOD (N00014-17-1-2175), the Bill and Melinda Gates Foundation, and the Cornell Center for Materials Research with funding from the NSF MRSEC program (DMR-1719875). We are thankful for generous support by Zillow and SAP America Inc.}
	\clearpage
	{\small
		\bibliographystyle{ieee_fullname}
		\bibliography{main}
	}
	\appendix
	\clearpage
	\newpage
	\begin{center}
	 \textbf{\Large Supplementary Material}
	\end{center}
	
	\input{supplementary}

\end{document}

%% file: math_definition.tex
\usepackage{amssymb}
\usepackage{amsmath,amsfonts}
\usepackage{amsopn}
\usepackage{bm} 
\usepackage{multirow}

\newcommand{\vct}[1]{\boldsymbol{#1}} 
\newcommand{\mat}[1]{\boldsymbol{#1}} 

\newcommand{\ProbOpr}[1]{\mathbb{#1}}

\newcommand{\expect}[2]{%
\ifthenelse{\equal{#2}{}}{\ProbOpr{E}_{#1}}
{\ifthenelse{\equal{#1}{}}{\ProbOpr{E}\left[#2\right]}{\ProbOpr{E}_{#1}\left[#2\right]}}} 
\newcommand{\var}[2]{%
\ifthenelse{\equal{#2}{}}{\ProbOpr{VAR}_{#1}}
{\ifthenelse{\equal{#1}{}}{\ProbOpr{VAR}\left[#2\right]}{\ProbOpr{VAR}_{#1}\left[#2\right]}}} 

\DeclareMathOperator*{\argmin}{\mathrm{argmin}}

\newcommand{\vp}{\vct{p}}

\newcommand{\mA}{\mat{A}}

\newcommand{\mT}{\mat{T}}

\newcommand{\eat}[1]{}
\newcommand{\method}[1]{\textsc{#1}}

\newcommand{\SDN}{\method{SDN}\xspace}

\newcommand{\vPIXOR}{\method{PIXOR$^\star$}\xspace}
\newcommand{\PIXOR}{\method{PIXOR}\xspace}
\newcommand{\PRCNN}{\method{P-RCNN}\xspace}
\newcommand{\SRCNN}{\method{S-RCNN}\xspace}
\newcommand{\PL}{\method{pseudo-LiDAR}\xspace}

\newcommand{\DOP}{\method{3DOP}\xspace}

\newcommand{\MLFstereo}{\method{MLF-stereo}\xspace}

\newcommand{\APBEV}{AP$_\text{BEV}$\xspace}
\newcommand{\AP}{AP$_\text{3D}$\xspace}

\newcommand{\RTD}{\method{RT3DStereo}\xspace}
\newcommand{\OCS}{\method{OC-Stereo}\xspace}
\newcommand{\ETE}{\method{E2E-PL}\xspace}

%% file: abs.tex
\begin{abstract}
	Reliable and accurate 3D object detection is a necessity for safe autonomous driving. Although LiDAR sensors can provide accurate 3D point cloud estimates of the environment, they are also prohibitively expensive for many settings. Recently, the introduction of \emph{pseudo-LiDAR} (PL) has led to a drastic reduction in the accuracy gap between methods based on LiDAR sensors and those based on cheap stereo cameras. PL combines state-of-the-art deep neural networks for 3D depth estimation with those for 3D object detection by converting 2D depth map outputs to 3D point cloud inputs. However, so far these two networks have to be trained separately. In this paper, we introduce a new framework based on differentiable Change of Representation (CoR) modules that allow the entire PL pipeline to be trained \emph{end-to-end}.  The resulting framework is compatible with most state-of-the-art networks for both tasks and in combination with PointRCNN improves over PL consistently across all benchmarks --- yielding the highest entry on the KITTI image-based 3D object detection leaderboard at the time of submission. Our code will be made available at \url{https://github.com/mileyan/pseudo-LiDAR_e2e}.
\end{abstract}

%% file: intro.tex
\section{Introduction}
\label{sec:intro}

\begin{figure}[t]
	\centering
	\small
	\centerline{\includegraphics[width=0.9\linewidth]{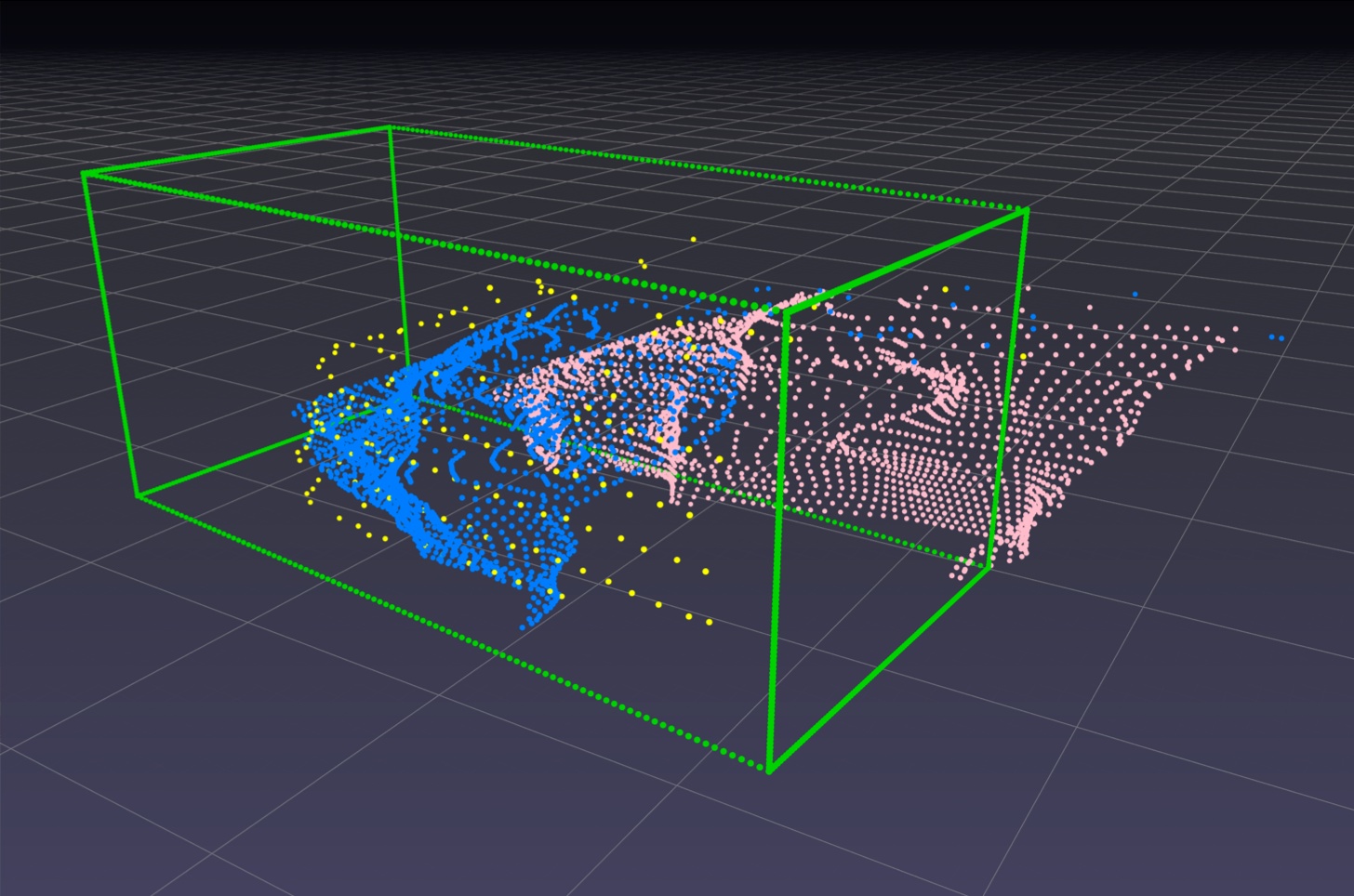}}
	\vskip-5pt
	\caption{\small \textbf{An illustration of the effectiveness of our end-to-end pipeline.} The green bounding box is the ground truth detection of a car. The yellow points are points from LiDAR. The pink point cloud is generated from an independently trained depth estimator, which is inaccurate and lies out of the green box. By making depth estimation and 3D object detection end-to-end, we obtain a better blue point cloud. Upon this, the object detector could yield the state-of-the-art performance.}
	\label{fig:1}
	\vskip-10pt
\end{figure}

One of the most critical components in autonomous driving is 3D object detection: a self-driving car must accurately detect and localize objects such as cars and pedestrians in order to plan the path safely and avoid collisions. To this end, existing algorithms primarily rely on LiDAR (Light Detection and Ranging) as the input signal, which provides precise 3D point clouds of the surrounding environment. LiDAR, however, is very expensive. A 64-beam model can easily cost more than the car alone, making self-driving cars prohibitively expensive for the general public. 

One solution is to explore alternative sensors like commodity (stereo) cameras. Although there is still a noticeable gap to LiDAR, it is an area with exceptional progress in the past year~\cite{konigshofrealtime,li2019stereo,pon2019object,pseudoLiDAR,xu2020zoomnet,you2019pseudo}. For example, pseudo-LiDAR (PL)~\cite{pseudoLiDAR,you2019pseudo} converts a depth map estimated from stereo images into a 3D point cloud, followed by applying (any) existing LiDAR-based detectors. Taking advantage of the state-of-the-art algorithms from both ends~\cite{chang2018pyramid,ku2018joint,qi2018frustum,shi2019pointrcnn,you2019pseudo}, pseudo-LiDAR achieves the highest image-based 3D detection accuracy ($34.1\%$ and $42.4\%$ at the moderate case) on the KITTI leaderboard~\cite{geiger2013vision,geiger2012we}.

While the modularity of pseudo-LiDAR is conceptual appealing, the combination of two independently trained components can yield an undesired performance hit. In particular, pseudo-LiDAR requires two systems: a depth estimator, typically trained on a generic depth estimation (stereo) image corpus, and an object detector trained on the point cloud data converted from the resulting depth estimates. 
It is unlikely that the two training objectives are optimally aligned for the ultimate goal, to maximize final detection accuracy. For example, depth estimators are typically trained with a loss that penalizes errors across all pixels equally, instead of focusing on objects of interest. Consequently, it may over-emphasize nearby or non-object pixels as they are over-represented in the data. Further, if the depth network is trained to estimate disparity, its intrinsic error will be exacerbated for far-away objects~\cite{you2019pseudo}. 

To address these issues, we propose to design a 3D object detection framework that is trained \emph{end-to-end}, while preserving the modularity and compatibility of pseudo-LiDAR with newly developed depth estimation and object detection algorithms. To enable back-propagation based end-to-end training on the final loss, the \emph{change of representation (CoR)} between the depth estimator and the object detector must be differentiable with respect to the estimated depth. We focus on two types of CoR modules --- subsampling and quantization --- which are compatible with different LiDAR-based object detector types. We study in detail on how to enable effective back-propagation with each module. 
Specifically, for quantization, we introduce a novel differentiable soft quantization CoR module to overcome its inherent non-differentiability. The resulting framework is readily compatible with most existing (and hopefully future) LiDAR-based detectors and 3D depth estimators.

We validate our proposed end-to-end pseudo-LiDAR (E2E-PL) approach with two representative object detectors --- \PIXOR~\cite{yang2018pixor} (quantized input) and PointRCNN~\cite{shi2019pointrcnn} (subsampled point input) --- on the widely-used KITTI object detection dataset~\cite{geiger2013vision,geiger2012we}. Our results are promising: we improve over the baseline pseudo-LiDAR pipeline and the improved PL++ pipeline~\cite{you2019pseudo} in all the evaluation settings and  significantly outperform other image-based 3D object detectors. 
At the time of submission our \ETE with PointRCNN holds the best results on the KITTI image-based 3D object detection leaderboard. Our qualitative results further confirm that end-to-end training can effectively guide the depth estimator to refine its estimates around object boundaries, which are crucial for accurately localizing objects (see \autoref{fig:1} for an illustration). 

%% file: related.tex
\section{Related Work}
\label{sec:related}
\noindent\textbf{3D Object Detection.}
Most works on 3D object detection are based on 3D LiDAR point clouds~\cite{chen2019fast,du2018general,engelcke2017vote3deep,lang2019pointpillars,li20173d,li2016vehicle,meyer2019lasernet,shi2019pointrcnn,shi2020points,yan2018second,yang2018hdnet,yang2019std}. Among these, there are two streams in terms of point cloud processing: 1) directly operating on the unordered point clouds in 3D \cite{lang2019pointpillars,qi2018frustum, shi2019pointrcnn, zhou2018voxelnet}, mostly by applying PointNet~\cite{qi2017pointnet,qi2017pointnet++} or/and applying 3D convolution over neighbors;
2) operating on quantized 3D/4D tensor data, which are generated from discretizing the locations of point clouds into some fixed grids~\cite{chen2017multi,ku2018joint,liang2018deep,yang2018pixor}.
Images can be included in both types of approaches, but primarily to supplement LiDAR signal~\cite{chen2017multi,du2018general,ku2018joint,liang2019multi,liang2018deep,meyer2019sensor,qi2018frustum,xu2018pointfusion}.

Besides LiDAR-based models, there are solely image-based models, which are mostly developed from the 2D frontal-view detection pipeline~\cite{he2017mask,lin2017feature,ren2015faster}, but most of them are no longer competitive with the state of the art in  localizing objects in 3D~\cite{chabot2017deep,chen20153d,chen20183d,chen2016monocular,mousavian20173d,li2019gs3d,pham2017robust,xiang2015data,xiang2017subcategory,xu2018multi}.

\noindent\textbf{Pseudo-LiDAR.}
This gap has been greatly reduced by the recently proposed pseudo-LiDAR framework~\cite{pseudoLiDAR, you2019pseudo}. Different from previous image-based 3D object detection models, pseudo-LiDAR first utilizes an image-based depth estimation model to obtain predicted depth $Z(u,v)$ of each image pixel $(u,v)$. The resulting depth $Z(u,v)$ is then projected to a ``pseudo-LiDAR'' point $(x, y, z)$ in 3D by
\begin{align}
	z = Z(u, v), \hspace{5pt}  x = \frac{(u - c_U)\cdot z}{f_U}, \hspace{5pt} y = \frac{(v - c_V)\cdot z}{f_V}, \label{eq_PLP}
\end{align}
where $(c_U, c_V)$ is the camera center and $f_U$ and $f_V$ are the horizontal and vertical focal lengths. The ``pseudo-LiDAR'' points are then treated as if they were LiDAR signals, over which any LiDAR-based 3D object detector can be applied. By making use of the separately trained state-of-the-art algorithms from both ends~\cite{chang2018pyramid, ku2018joint,qi2018frustum, shi2019pointrcnn,you2019pseudo}, pseudo-LiDAR achieved the highest image-based performance on KITTI benchmark~\cite{geiger2013vision,geiger2012we}.
Our work builds upon this framework.

%% file: approach.tex
\section{End-to-End Pseudo-LiDAR}
\label{sec:approach}

One key advantage of the pseudo-LiDAR pipeline~\cite{pseudoLiDAR, you2019pseudo} is its \emph{plug-and-play} modularity, which allows it to incorporate any advances in 3D depth estimation or LiDAR-based 3D object detection. However, it also lacks the notion of end-to-end training of both components to ultimately maximize the detection accuracy. In particular, the pseudo-LiDAR pipeline is trained in two steps, with different objectives. First, a depth estimator is learned to estimate generic depths for all pixels in a stereo image; then a LiDAR-based detector is trained to predict object bounding boxes from depth estimates, generated by the frozen depth network. 

\begin{figure}[t]
	\begin{center}
		\includegraphics[width=.9 \linewidth]{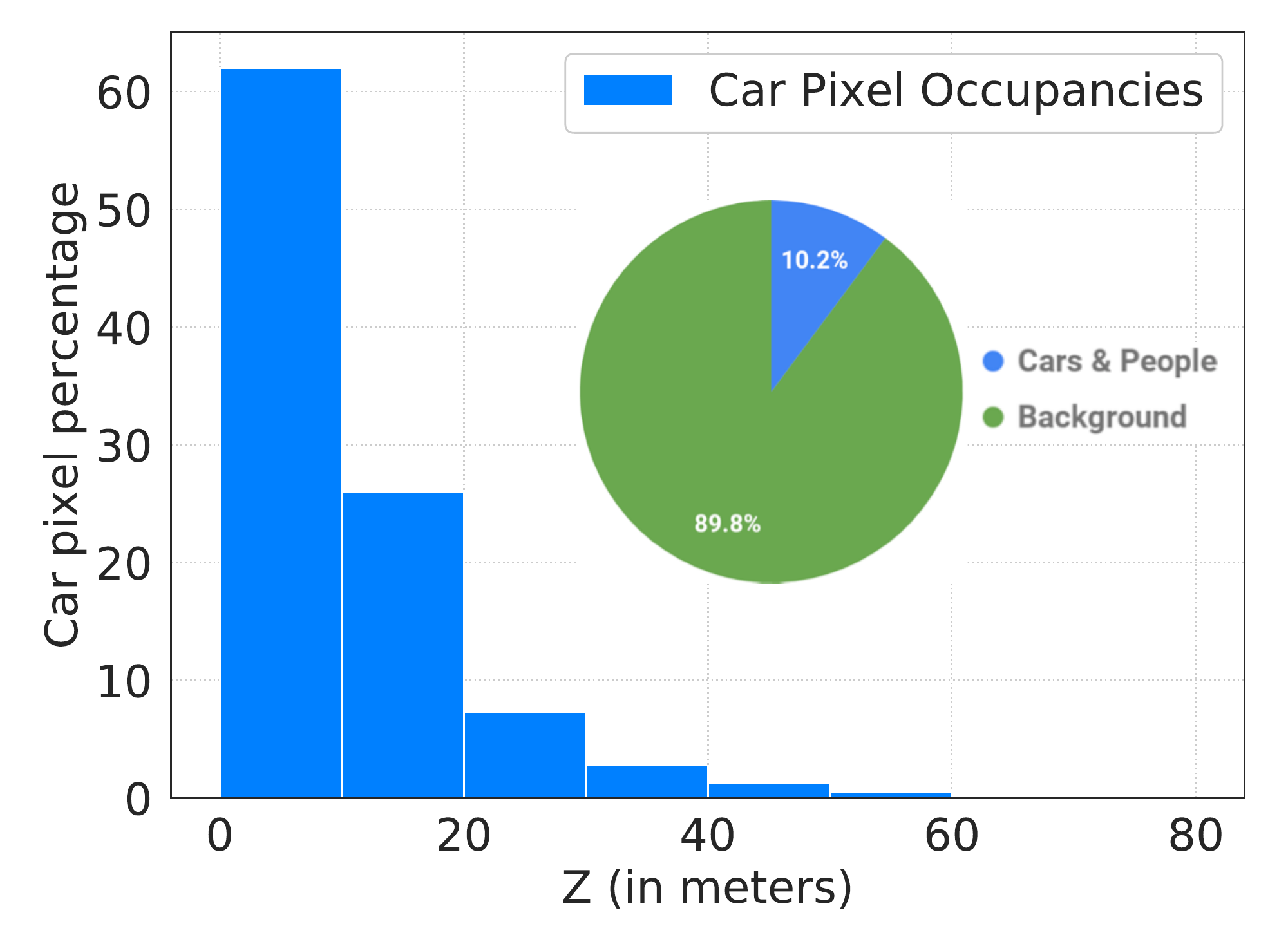}
	\end{center}
	\vskip-20pt
	\caption{\small\textbf{Pixel distribution:} $90\%$ of all pixels correspond to background. The 10\% pixels associated with cars and people ($<1$\% people) are primarily within a depth of 20m.}
	\label{fg:car_occ}
	\vskip-10pt
\end{figure}

As mentioned in \autoref{sec:intro}, learning pseudo-LiDAR in this fashion does not align the two components well. On one end, a LiDAR-based object detector heavily relies on accurate 3D points on or in the proximity of the object surfaces to detect and localize objects. Especially, for far-away objects that are rendered by relatively few points. 
On the other end, a depth estimator learned to predict all the pixel depths may place over-emphasis on the background and nearby objects since they occupy most of the pixels in an image. For example, in the KITTI dataset~\cite{jeong2019complex} only about $10\%$ of all pixels correspond to cars and pedestrians/cyclists (\autoref{fg:car_occ}). Such a misalignment is aggravated with fixing the depth estimator in training the object detector: the object detector is unaware of the intrinsic depth error in the input and thus can hardly detect the far-away objects correctly.

\begin{figure}[t]
	\centering
	\small
	\includegraphics[width=\linewidth]{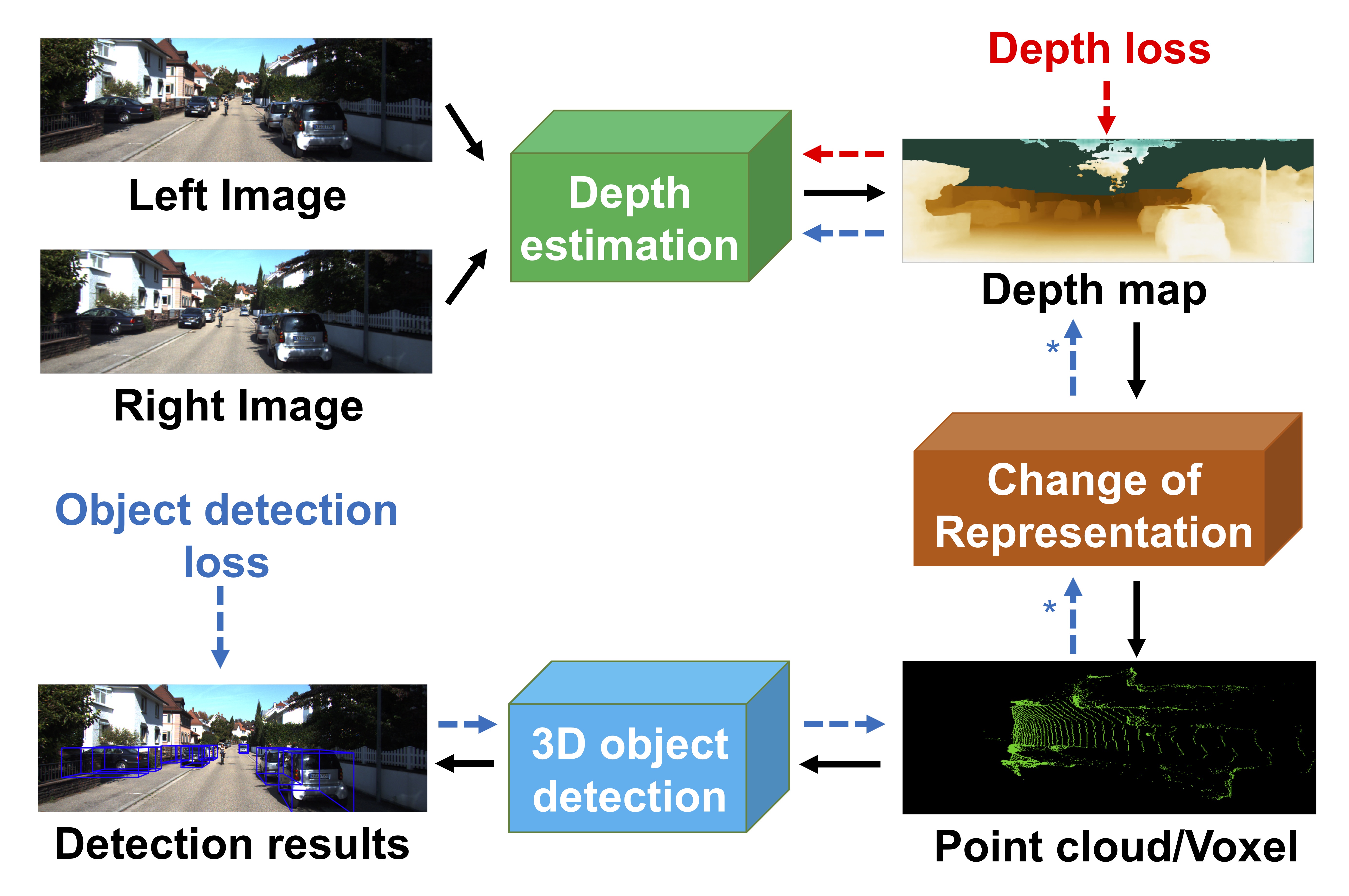}
	\vskip-10pt
	\caption{\small\textbf{End-to-end image-based 3D object detection:} We introduce a \emph{change of representation (CoR)} layer to connect the output of the depth estimation network as the input to the 3D object detection network. The result is an end-to-end pipeline that yields object bounding boxes directly from stereo images and allows back-propagation throughout all layers. 
		Black solid arrows represent the forward pass; {\color{blue}Blue} and {\color{red}red} dashed arrows represent the backward pass for the object detection loss and depth loss, respectively. The * denotes that our \emph{CoR} layer is able to back propogate the gradients between different representations.}
	\label{fig:e2e}
	\vskip-10pt
\end{figure}

\begin{figure*}[ht!]
	\centering
	\small
	\centerline{\includegraphics[width=0.95\linewidth]{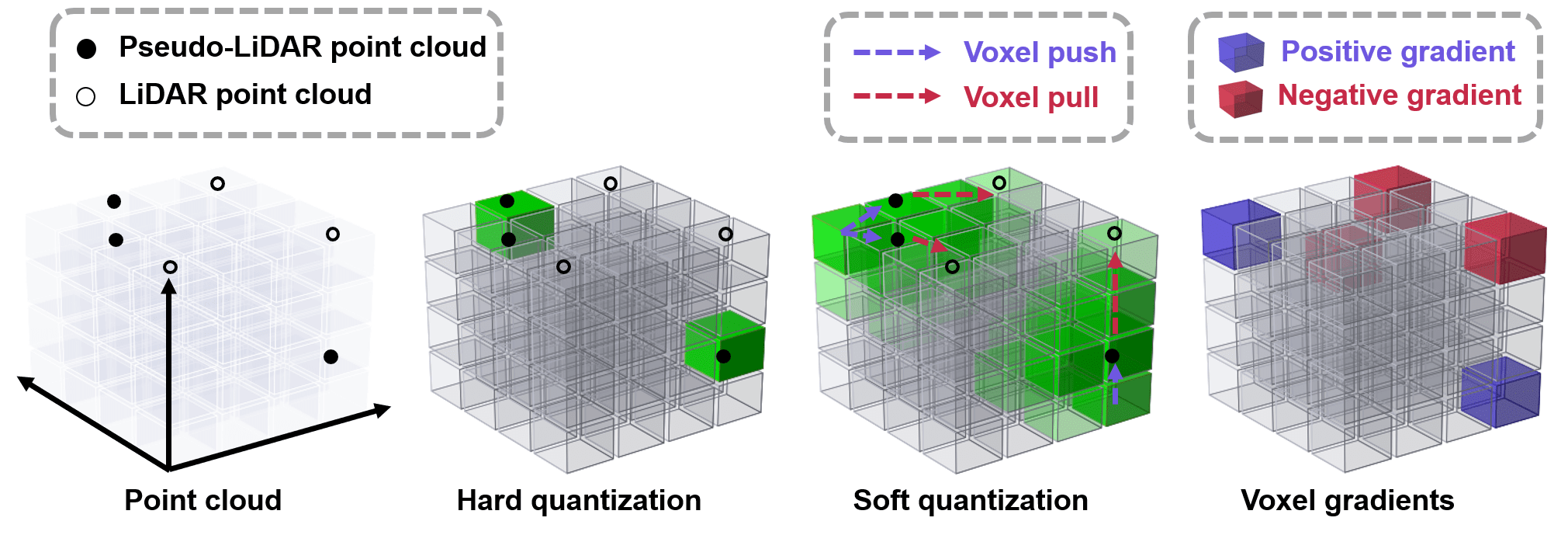}}
	\vskip-10pt
	\caption{\small\textbf{Quantization:} We voxelize an input pseudo-LiDAR (PL) point cloud using soft or hard quantization. Green voxels are those influenced by the PL points. 
		A blue voxel having a \textbf{positive} gradient of the detection loss $\mathcal{L}_\text{det}$ exerts a force to push points away from its center to other voxels, whereas a red voxel having a negative \textbf{gradient} exerts a force to pull points of other voxels to its center. These forces at the red and blue voxles can only affect PL points if PL points influence those voxels. Soft quantization increases the area of influence of the PL points and therefore the forces, allowing points from other voxels to be pushed away or pulled towards. The updated PL points thus can become closer to the ground truth LiDAR point cloud. 
	}
	\label{fig:quantization}
	\vskip-10pt
\end{figure*}

\autoref{fig:e2e} illustrates our proposed end-to-end pipeline to resolve these shortcomings. Here, the error signal from misdetecting or mislocalizing an object can ``softly attend'' to pixels which affect the prediction most (likely those on or around objects in 2D), instructing the depth estimator where to improve for the subsequent detector. To enable back-propagating the error signal from the final detection loss, the \emph{change of representation (CoR)} between the depth estimator and the object detector must be differentiable with respect to the estimated depth. In the following, we identify two major types of CoR --- subsampling and quantization --- in incorporating existing LiDAR-based detectors into the pseudo-LiDAR pipeline. 

\input{voxel_pipeline}

\input{pointcloud_pipeline}

\input{loss}

%% file: voxel_pipeline.tex
\subsection{Quantization}
\label{ssec:q}
Several LiDAR-based object detectors take voxelized 3D or 4D tensors as inputs~\cite{chen2017multi,ku2018joint,liang2018deep,yang2018pixor}. The 3D point locations are discretized into a fixed grid, and only the occupation (\ie, $\{0,1\}$) or densities (i.e., $[0,1]$) are recorded in the resulting tensor\footnote{For LiDAR data the reflection intensity is often also recorded.}. The advantage of this kind of approaches is that 2D and 3D convolutions can be directly applied to extract features from the tensor. Such a discretization process, however, makes the back-propagation difficult.

Let us consider an example where we are given a point cloud $\bm{P} = \{\bm{p}_1, \dots, \bm{p}_N\}$ with the goal to generate a 3D occupation tensor $\mT$ of $M$ bins, where each bin $m\in\{1,\cdots,M\}$ is associated with a fixed center location $\hat{\vp}_m$. The resulting tensor $T$ is defined as follows,
\begin{align}
	\mT(m)\! =\! \begin{cases}
		1, & \!\!\!\text{if } \exists~\vp\!\in\!\bm{P} \text{ s.t. } m\! =\! \underset{\ m'}{\argmin} \|\bm{p}-\hat{\vp}_{m'}\|_2 \\
		0, & \!\!\!\text{otherwise}.
	\end{cases} \label{eq_hard}
\end{align}
In other words, if a point $\bm{p}\in\bm{P}$ falls into bin $m$, then $\mT(m) = 1$; otherwise, $0$. 
The forward pass of generating $\mT$ is as straightforward. The backward pass to obtain the gradient signal of the detection loss $\mathcal{L}_\text{det}$ with respect to $\vp\in\bm{P}$ or the depth map $Z$ (\autoref{eq_PLP}), however, is non-trivial.

Concretely, we can obtain $\nabla_{\mT}\mathcal{L}_\text{det}$ by taking the gradients of  $\mathcal{L}_\text{det}$ with respect to ${\mT}$. Intuitively, if $\frac{\partial\mathcal{L}_\text{det}}{\partial{\mT}(m)}<0$, it means that $\mT(m)$ should increase; \ie, there should be points falling into bin $m$. In contrast, if $\frac{\partial\mathcal{L}_\text{det}}{\partial{\mT}(m)}>0$, it means that $\mT(m)$ should decrease by pushing points out from bin $m$. But how can we pass these messages back to the input point cloud $\bm{P}$? More specifically, how can we translate the single digit $\frac{\partial\mathcal{L}_\text{det}}{\partial{\mT}(m)}$ of each bin to be useful information in 3D in order to adjust the point cloud $\bm{P}$?

As a remedy, we propose to modify the forward pass by introducing a differentiable soft quantization module (see \autoref{fig:quantization}). We introduce a radial basis function (RBF) around the center $\hat{\vp}_m$ of a given bin $m$. Instead of binary occupancy counters\footnote{We note that the issue of back-propagation cannot be resolved simply by computing real-value densities in~\autoref{eq_hard}.}, we keep a ``soft'' count of the points inside the bin, weighted by the RBF. Further, we allow any given bin $m$ to be influenced by a local neighborhood $\mathcal{N}_m$ of close bins. We then modify the definition of $\mT$ accordingly. 
Let $\bm{P}_m$ denote the set of points that fall into bin $m$,
\begin{align}
\bm{P}_m = \{\vp\in\bm{P}, \text{s.t. } m=\argmin_{m'} \|\bm{p}-\hat{\vp}_{m'}\|_2\}.\nonumber
\end{align}
We define $\mT(m, m')$ to denote the average RBF weight of points in bin $m'$ w.r.t. bin $m$ (more specifically, $\hat{\vp}_m$),   
\begin{align}
\!\!\mT(m, m') \!=\! 
\left\{\!\!
\begin{array}{cc}
0 & \textrm{if $|P_{m'}|=0$;} \\
\!\!\frac{1}{|\bm{P}_{m'}|} \underset{\vp\in\bm{P}_{m'}}{\sum} e^{-\frac{\|\bm{p} - \hat{\vp}_{m}\|^2}{\sigma^2} } & \textrm{if $|P_{m'}|>0$.}\label{eq_diff}\\
\end{array}\right.
\end{align} 
The final value of the tensor $\mT$ at bin $m$ is the combination of soft occupation from its own and neighboring bins,
\begin{align}
\mT(m) = 
\mT(m,m)+\frac{1}{|\mathcal{N}_m|}\sum_{m'\in \mathcal{N}_m} \mT(m, m'). \label{eqcomb}
\end{align}  
We note that, when $\sigma^2\!\gg\! 0$ and $\mathcal{N}_m=\varnothing$,  \autoref{eqcomb} recovers \autoref{eq_hard}. 
Throughout this paper, we set the neighborhood $\mathcal{N}_m$ to the 26 neighboring bins (considering a 3x3x3 cube centered on the bin) and $\sigma^2=0.01$. Following $\cite{yang2018pixor}$, we set the total number of bins to $M=700\times 800\times 35$.

Our soft quantization module is fully  differentiable. The partial derivative $\frac{\partial\mathcal{L}_\text{det}}{\partial{\mT}(m)}$ directly affects the points in bin $m$ (\ie, $\bm{P}_m$) and its neighboring bins and enables end-to-end training. For example, to pass the partial derivative to a point $\bm{p}$ in bin $m'$, we compute $\frac{\partial\mathcal{L}_\text{det}}{\partial{\mT}(m)} \times \frac{\partial{\mT}(m)}{\partial{\mT}(m, m')} \times \nabla_{\vp} \mT(m, m')$. More importantly, even when bin $m$ mistakenly contains no point, $\frac{\partial\mathcal{L}_\text{det}}{\partial{\mT}(m)}>0$ allows it to drag points from other bins, say bin $m'$, to be closer to $\hat{\vp}_m$, enabling corrections of the depth error more effectively.

%% file: pointcloud_pipeline.tex
\subsection{Subsampling}
\label{ssec:p}
As an alternative to voxelization, some LiDAR-based object detectors take the raw 3D points as input (either as a whole~\cite{shi2019pointrcnn} or by grouping them according to metric locations~\cite{lang2019pointpillars,yan2018second,zhou2018voxelnet} or potential object locations~\cite{qi2018frustum}). For these, we can directly  use the 3D point clouds obtained by \autoref{eq_PLP}; however, some subsampling is required. Different from voxelization, subsampling is far more amenable to end-to-end training: the points that are filtered out can simply be ignored during the backwards pass; the points that are kept are left untouched.
First, we remove all 3D points higher than the normal heights that LiDAR signals can cover, such as pixels of the sky.
Further, we may sparsify the remaining points by subsampling. This second step is optional but suggested in~\cite{you2019pseudo} due to the significantly larger amount of points generated from depth maps than LiDAR: on average 300,000 points are in the pseudo-LiDAR signal but 18,000 points are in the LiDAR signal (in the frontal view of the car). Although denser representations can be advantageous in terms of accuracy, they do slow down the object detection network. 
We apply an angular-based sparsifying method. We define multiple bins in 3D by discretizing the spherical coordinates $(r, \theta, \phi)$. Specifically, we discretize $\theta$ (polar angle) and $\phi$ (azimuthal angle) to mimic the LiDAR beams. We then keep a single 3D point $(x, y, z)$ from those points whose spherical coordinates fall into the same bin. The resulting point cloud therefore mimics true LiDAR points.

In terms of back-propagation, since these 3D object detectors directly process the 3D coordinates $(x,y,z)$ of a point, we can obtain the gradients of the final detection loss $\mathcal{L}_\text{det}$ with respect to the coordinates; i.e., $(\frac{\partial\mathcal{L}_\text{det}}{\partial x},\frac{\partial\mathcal{L}_\text{det}}{\partial y},\frac{\partial\mathcal{L}_\text{det}}{\partial z})$. As long as we properly record which points are subsampled in the forward pass or how they are grouped, back-propagating the gradients from the object detector to the depth estimates $Z$ (at sparse pixel locations) can be straightforward. Here, we leverage the fact that~\autoref{eq_PLP} is differentiable with respect to $z$. However, due to the high  sparsity of gradient information in $\nabla_Z \mathcal{L}_\text{det}$, we found that the initial depth loss used to train a conventional depth estimator is required to jointly optimize the depth estimator. 

This subsection, together with~\autoref{ssec:q}, presents a general end-to-end framework applicable 
to various object detectors. We do not claim this subsection as a technical contribution, but it provides details that makes end-to-end training for point-cloud-based detectors successful.

%% file: loss.tex
\subsection{Loss}
To learn the pseudo-LiDAR framework end-to-end, we replace
\autoref{eq_hard} by \autoref{eqcomb} for object detectors that take 3D or 4D tensors as input. For object detectors that take raw points as input, no specific modification is needed.

We learn the object detector and the depth estimator jointly with the following loss,  
\begin{displaymath}
\mathcal{L} = \lambda_\text{det} \mathcal{L}_\text{det} + \lambda_\text{depth} \mathcal{L}_\text{depth},
\label{eq_total_loss}
\end{displaymath}
where $\mathcal{L}_\text{det}$ is the loss from 3D object detection and $\mathcal{L}_\text{depth}$ is the loss of depth estimation. 
$\lambda_\text{depth}$ and $\lambda_\text{det}$ are the corresponding coefficients.
The detection loss $\mathcal{L}_\text{det}$ is the combination of classification loss and regression loss,
\begin{displaymath}
\mathcal{L}_\text{det} = \lambda_\text{cls} \mathcal{L}_\text{cls} + \lambda_\text{reg} \mathcal{L}_\text{reg},
\label{eq_det_loss}
\end{displaymath}
in which the classification loss aims to assign correct class (e.g., car) to the detected bounding box; the regression loss aims to refine the size, center, and rotation of the box.

Let $Z$ be the predicted depth and $Z^*$ be the ground truth, we apply the following depth estimation loss
\begin{displaymath}
\mathcal{L}_\text{depth} = \frac{1}{|\mA|} \sum_{(u, v)\in\mA} \ell(Z(u, v) - Z^*(u, v)),
\label{eq_depth_loss}
\end{displaymath}
where $\mA$ is the set of pixels that have ground truth depth. $\ell(x)$ is the smooth L1 loss defined as
\begin{align}
\ell(x) =
\begin{cases}
0.5 x^2, \text{ if } |x|<1;\\
|x|-0.5, \text{ otherwise.}
\end{cases}
\end{align}
We find that the depth loss is important as the loss from object detection may only influence parts of the pixels (due to quantization or subsampling).  
After all, our hope is to make the depth estimates around (far-away) objects more accurately, but not to sacrifice the accuracy of depths on the background and nearby objects\footnote{The depth loss can be seen as a regularizer to keep the output of the depth estimator physically meaningful. We note that, 3D object detectors are designed with an inductive bias: the input is an accurate 3D point cloud. However, with the large capacity of neural networks, training the depth estimator and object detector end-to-end with the detection loss alone can lead to arbitrary representations between them that break the inductive bias but achieve a lower training loss. The resulting model thus will have a much worse testing loss than the one trained together with the depth loss.}. 

%% file: exp.tex
\section{Experiments}
\label{sec:exp}

\subsection{Setup}
\label{ssec:setup}

\begin{table}[!tbp]
\centering
\caption{\small {Statistics of the gradients of different losses on the predicted depth map. Ratio: the percentage of pixels with gradients.}}
\vskip -5pt
\begin{tabular}{c|c|c|c}
      & Depth Loss & \PRCNN Loss & \PIXOR Loss \\ \hline
\hline Ratio & 3\%  & 4\%  & 70\%        \\ \hline
Mean  & $10^{-5}$ & $10^{-3}$  & $10^{-5}$             \\ \hline
Sum   & 0.1 & 10  &    1         \\ \hline
\end{tabular}
\label{tbGrad}
\vskip -5pt
\end{table}

\begin{table*}[!th]
	\centering
	\caption{\small \textbf{3D object detection results on the KITTI validation set.} We report \APBEV ~/ \AP (in \%) of the \textbf{car} category, corresponding to average precision of the bird's-eye view and 3D object detection. We arrange methods according to the input signals: S for stereo images, L for 64-beam LiDAR, M for monocular images. PL stands for \PL. \emph{Results of our end-to-end \PL are in {\color{blue} blue}.} Methods with 64-beam LiDAR are in {\color{gray} gray}. Best viewed in color.} \label{tbMain}
	\vskip-5pt
	\begin{tabular}{=l|+c|+c|+c|+c|+c|+c|+c}
		&  & \multicolumn{3}{c|}{IoU = 0.5} & \multicolumn{3}{c}{IoU = 0.7} \\ \cline{3-8}
		\multicolumn{1}{c|}{Detection algo} & Input & Easy & Moderate & Hard & Easy & Moderate & Hard \\ \hline
		\DOP~\cite{chen20153d} & S & 55.0 / 46.0 & 41.3 / 34.6 & 34.6 / 30.1 & 12.6 / 6.6 \hspace{3pt} & 9.5 / 5.1 & 7.6 / 4.1 \\
		\MLFstereo~\cite{xu2018multi} & S & - & 53.7 / 47.4 & - & - & 19.5 / 9.8 \hspace{3pt} & - \\
		\SRCNN~\cite{li2019stereo} & S & 87.1 / 85.8 & 74.1 / 66.3  & 58.9  / 57.2 & 68.5 / 54.1 & 48.3 / 36.7 &  41.5 / 31.1 \\
		\OCS~\cite{pon2019object} & S & 90.0 / 89.7 & 80.6 / 80.0 & 71.1 / 70.3 & 77.7 / 64.1 & 66.0 / 48.3 & 51.2 / 40.4 \\ \hline
		PL: \PRCNN~\cite{pseudoLiDAR} & S & 88.4 / 88.0 & 76.6 / 73.7 & 69.0 / 67.8 &  73.4 / 62.3 & 56.0 / 44.9 & 52.7 / 41.6 \\
		PL++: \PRCNN~\cite{you2019pseudo}  & S & 89.8 / {89.7} & 83.8 / 78.6 & {77.5} / {75.1} & 82.0 / 67.9 & 64.0 / 50.1 & 57.3 / 45.3 \\
		\rowstyle{\color{blue}}
		\hspace{-3pt}\ETE: \PRCNN & S & \textbf{90.5} / \textbf{90.4}  & \textbf{84.4} / \textbf{79.2}  & \textbf{78.4} / \textbf{75.9} & \textbf{82.7} / \textbf{71.1} & \textbf{65.7} / \textbf{51.7} & \textbf{58.4} / \textbf{46.7}  \\ \hline

		PL: \vPIXOR~\cite{pseudoLiDAR} & S & 89.0 / - \hspace{12pt} & 75.2 / - \hspace{12pt} & 67.3 / - \hspace{12pt} & 73.9 / - \hspace{12pt} & 54.0 / - \hspace{12pt} & 46.9 / - \hspace{12pt} \\
		PL++: \vPIXOR~\cite{you2019pseudo}  & S & {89.9} / - \hspace{12pt} & 78.4 / - \hspace{12pt} & 74.7 / - \hspace{12pt} & 79.7 / - \hspace{12pt}   & 61.1 / - \hspace{12pt}  & 54.5 / - \hspace{12pt} \\
        \rowstyle{\color{blue}}
		\hspace{-3pt}\ETE: \vPIXOR  & S & \textbf{94.6} / - \hspace{12pt} & \textbf{84.8} / - \hspace{12pt} &  \textbf{77.1}/ - \hspace{12pt} & \textbf{80.4} / - \hspace{12pt} & \textbf{64.3} / - \hspace{12pt} & \textbf{56.7} / - \hspace{12pt} \\
		\hline
		\rowstyle{\color{gray}}
		\hspace{-5pt}\PRCNN~\cite{shi2019pointrcnn} & L  & 97.3 / 97.3 & 89.9 / 89.8 & 89.4 / 89.3 &  90.2 / 89.2 & 87.9 / 78.9 & 85.5 / 77.9 \\
		\rowstyle{\color{gray}}
		\hspace{-2.5pt}\vPIXOR~\cite{yang2018pixor} & L + M & 94.2 / - \hspace{12pt} & 86.7 / - \hspace{12pt} & 86.1 / - \hspace{12pt} & 85.2 / - \hspace{12pt} & 81.2 / - \hspace{12pt} & 76.1 / - \hspace{12pt} \\
		\hline
	\end{tabular}
	\vskip-5pt
\end{table*}

\noindent\textbf{Dataset.}
We evaluate our end-to-end (\ETE) approach on the KITTI object detection benchmark~\cite{geiger2013vision,geiger2012we}, which contains 3,712, 3,769 and 7,518 images for training, validation, and testing. KITTI provides for each image the corresponding 64-beam Velodyne LiDAR point cloud, right image for stereo, and camera calibration matrices.

\noindent\textbf{Metric.}
We focus on 3D and bird's-eye-view (BEV) object detection and report the results on the \emph{validation set}. We focus on the ``car'' category, following~\cite{chen2017multi,pseudoLiDAR, xu2018pointfusion}. We report the average precision (AP) with the IoU thresholds at 0.5 and 0.7. We denote AP for the 3D and BEV tasks by \AP and \APBEV. KITTI defines the easy, moderate, and hard settings, in which objects with 2D box heights smaller than or occlusion/truncation levels larger than certain thresholds are disregarded. The hard (moderate) setting contains all the objects in the moderate and easy (easy) settings.

\noindent\textbf{Baselines.}
We compare to seven stereo-based 3D object detectors: \PL~(PL)~\cite{pseudoLiDAR}, \PL++~(PL++)~\cite{you2019pseudo}, \DOP~\cite{chen20153d}, \SRCNN~\cite{li2019stereo}, \RTD~\cite{konigshofrealtime}, \OCS~\cite{pon2019object}, and \MLFstereo~\cite{xu2018multi}. \emph{For \PL++, we only compare to its image-only method.}

\subsection{Details of our approach}
\label{exp_approach}
Our end-to-end pipeline has two parts: stereo depth estimation and 3D object detection. In training, we first learn only the stereo depth estimation network to get a depth estimation prior, and then we fix the depth network and use its output to train the 3D object detector from scratch. In the end, we joint train the two parts with balanced loss weights.

\noindent\textbf{Depth estimation.}
We apply \SDN~\cite{you2019pseudo} as the backbone to estimate a dense depth map $Z$. We follow~\cite{you2019pseudo} to pre-train \SDN{} on the synthetic Scene Flow dataset~\cite{mayer2016large}
and fine-tune it on the 3,712 training images of KITTI. We obtain the depth ground truth $Z^*$ by projecting the corresponding LiDAR points onto images.

\noindent\textbf{Object detection.} We apply two LiDAR-based algorithms: \PIXOR~\cite{yang2018pixor} (voxel-based, with quantization) and PointRCNN (\PRCNN)~\cite{shi2019pointrcnn} (point-cloud-based).
We use the released code of \PRCNN. We obtain the code of \PIXOR from the authors of~\cite{you2019pseudo}, which has slight modification to include visual information (denoted as \vPIXOR). 

\noindent\textbf{Joint training.} We set the depth estimation and object detection networks trainable, and allow the gradients of the detection loss to back-propagate to the depth network.
We study the gradients of the detection and depth losses w.r.t. the predicted depth map $Z$ to determine the hyper-parameters $\lambda_\text{depth}$ and $\lambda_\text{det}$.
For each loss, we calculate the percentage of pixels on the entire depth map that have gradients. We further collect the mean and sum of the gradients on the depth map during training, as shown in \autoref{tbGrad}.
The depth loss only influences 3\% of the depth map because the ground truth obtained from LiDAR is sparse. The \PRCNN loss, due to subsampling on the dense PL point cloud, can only influence 4\% of the depth map. For the \PIXOR loss, our soft quantization module could back-propagate the gradients to 70\% of the pixels of the depth map. In our experiments, we find that balancing the sums of gradients between the detection and depth losses is crucial in making joint training stable.
We carefully set $\lambda_\text{depth}$ and $\lambda_\text{det}$ to make sure that the sums are on the same scale in the beginning of training. For \PRCNN, we set $\lambda_\text{depth} = 1$ and $\lambda_\text{det} = 0.01$; for \PIXOR, we $\lambda_\text{depth} = 1$ and $\lambda_\text{det} = 0.1$.

\subsection{Results}
\label{ssec:results}

\noindent\textbf{On KITTI validation set.} The main results on KITTI validation set are summarized in \autoref{tbMain}. It can be seen that \textbf{1)} the proposed \ETE framework consistently improves the object detection performance on both the model using subsampled point inputs (\PRCNN) and that using quantized inputs (\vPIXOR). \textbf{2)} While the quantization-based model (\vPIXOR) performs worse than the point-cloud-based model (\PRCNN) when they are trained in a non end-to-end manner, end-to-end training greatly reduces the performance gap between these two types of models, especially for IoU at $0.5$: the gap between these two models on \APBEV in moderate cases is reduced from $5.4\%$ to $-0.4\%$. As shown in \autoref{tbGrad}, on depth maps, the gradients flowing from the loss of the \vPIXOR detector are much denser than those from the loss of the \PRCNN detector, suggesting that more gradient information is beneficial. \textbf{3)} For IoU at 0.5 under easy and moderate cases, \ETE: \vPIXOR performs on a par with \vPIXOR using LiDAR.

\begin{table}[!htb]
\tabcolsep 2pt
\centering
\caption{\small \textbf{3D object (car) detection results on the KITTI test set.} We compare \ETE ({\color{blue}blue}) with existing results retrieved from the KITTI leaderboard, and report \APBEV~/ \AP at IoU=0.7.}
\vskip -5pt
\begin{tabular}{=l|+c|+c|+c}
	Method & Easy & Moderate & Hard \\ \hline
	\SRCNN~\cite{li2019stereo} & 61.9 / 47.6  & 41.3 / 30.2& 33.4 / 23.7 \\
	\RTD~\cite{konigshofrealtime} & 58.8 / 29.9  & 46.8 / 23.3& 38.4 / 19.0 \\
	\OCS~\cite{pon2019object} &	68.9 / 55.2  & 51.5 /  37.6& 	43.0 / 30.3 \\
	PL~\cite{pseudoLiDAR} & 67.3 / 54.5  & 45.0 / 34.1 & 38.4 / 28.3 \\
	PL++: \PRCNN~\cite{you2019pseudo} & 78.3 / 61.1  &  58.0 / 42.4  & 51.3 / 37.0 \\
	\rowstyle{\color{blue}}
	\hspace{-3pt}\ETE: \PRCNN &  \textbf{79.6} / \textbf{64.8} &  \textbf{58.8} / \textbf{43.9} & \textbf{52.1} / \textbf{38.1} \\
	\hline
	PL++:\vPIXOR~\cite{you2019pseudo} & 70.7 / - \hspace{12pt}  & 48.3 / - \hspace{12pt} & 41.0  / - \hspace{12pt} \\
	\rowstyle{\color{blue}}
	\hspace{-3pt}\ETE: \vPIXOR &  \textbf{71.9} / - \hspace{12pt} & \textbf{51.7} / - \hspace{12pt} &  \textbf{43.3} / - \hspace{12pt} \\
	\hline
\end{tabular}
\label{tbTest}
\vskip-10pt
\end{table}

\noindent\textbf{On KITTI test set.} \autoref{tbTest} shows the results on KITTI \emph{test} set. We observe the same consistent performance boost by applying our \ETE framework on each detector type. At the time of submission, \ETE: \PRCNN achieves the state-of-the-art results over image-based models.

\begin{table*}[t]
	\centering
	\caption{\small \textbf{Ablation studies on the point-cloud-based pipeline with \PRCNN.} We report \APBEV ~/ \AP (in \%) of the \textbf{car} category, corresponding to average precision of the bird's-eye view and 3D detection. We divide our pipeline with \PRCNN into three sub networks: Depth, RPN and RCNN. $\surd$ means that we set the sub network trainable and use its corresponding loss in joint training. We note that the gradients of the later sub network would also back-propagate to the previous sub network. For example, if we choose Depth and RPN, the gradients of RPN would also be back-propogated to the Depth network. The best result per column is in {\color{blue} blue}. Best viewed in color.}
	\vskip-5pt 
	\begin{tabular}{=l|+c|+c|+c|+c|+c|+c|+c|+c}
		& & & \multicolumn{3}{c|}{IoU = 0.5} & \multicolumn{3}{c}{IoU = 0.7} \\ \cline{3-8}
		\hline Depth & RPN & RCNN & Easy & Moderate & Hard & Easy & Moderate & Hard \\ \hline
		\hline & & & 89.8/89.7 & 83.8/78.6 & 77.5/75.1 & 82.0/67.9 & 64.0/50.1 & 57.3/45.3 \\
		\hline $\surd$ & & & 89.7/89.5 & 83.6/78.5 & 77.4/74.9 & 82.2/67.8 & 64.5/50.5 & 57.4/45.4 \\
		\hline  & $\surd$ & & 89.3/89.0 & 83.7/78.3 & 77.5/75.0 & 81.1/66.5 & 63.9/50.0 & 57.1/45.2 \\
		\hline  & & $\surd$ & 89.6/89.4 & 83.9/78.2 & 77.6/75.2 & 81.7/68.2 & 63.4/50.4 & 57.2/45.9 \\
		\hline & $\surd$ & $\surd$ & 90.2/90.1 & 84.2/78.8 & 78.0/75.7 & 81.9/69.1 & 64.0/51.2 & 57.7/46.1 \\
		\hline $\surd$ & $\surd$ & & 89.3/89.1 & 83.9/78.5 & 77.7/75.2 & 81.3/69.4 & 64.7/50.7 & 57.7/45.7 \\
		\hline $\surd$ & & $\surd$ & 89.8/89.7 & 84.2/79.1 & 78.2/{\color{blue}76.5} & {\color{blue}84.2}/69.9 & 65.5/51.0 & 58.1/46.2 \\
		\hline $\surd$ & $\surd$ & $\surd$ & {\color{blue}90.5}/{\color{blue}90.4} & {\color{blue}84.4}/{\color{blue}79.2} & {\color{blue}78.4}/{75.9} & {82.7}/{\color{blue}71.1} & {\color{blue}65.7}/{\color{blue}51.7} & {\color{blue}58.4}/{\color{blue}46.7} \\ \hline
	\end{tabular}
	\label{tbAblation}
\end{table*}

\subsection{Ablation studies}
We conduct ablation studies on the the point-cloud-based pipeline with \PRCNN in \autoref{tbAblation}. We divide the pipeline into three sub networks: depth estimation network (Depth), region proposal network (RPN) and Regional-CNN (RCNN).
We try various combinations of the sub networks (and their corresponding losses) by setting them to trainable in our final joint training stage. The first row serves as the baseline. In rows two to four, the results indicate that simply training each sub network independently with more iterations does not improve the accuracy. In row five, joint training RPN and RCNN (\ie, \PRCNN) does not have significant improvement, because the point cloud from Depth is not updated and remains noisy.
In row six we jointly train Depth with RPN, but the result does not improve much either. We suspect that the loss on RPN is insufficient to guide the refinement of depth estimation. By combining the three sub networks together and using the RCNN, RPN, and Depth losses to refine the three sub networks, we get the best results (except for two cases).

For the quantization-based pipeline with soft quantization, we also conduct similar ablation studies, as shown in \autoref{tbAblation_PIXOR}. Since \vPIXOR is a one-stage detector, we divide the pipeline into two components: Depth and Detector. Similar to the point-cloud-based pipeline (\autoref{tbAblation}), simply training each component independently with more iterations does not improve. However, when we jointly train both components, we see a significant improvement (last row). 
This demonstrates the effectiveness of our soft quantization module which can back-propagate the Detector loss to influence 70\% pixels on the predicted depth map.

More interestingly, applying the \emph{soft quantization} module
alone without jointly training (rows one and two) does not improve over or is even outperformed by PL++:
\vPIXOR with \emph{hard quantization} (whose result is $89.9 / 79.7 | 78.4 / 61.1 | 74.7 /54.5$, from~\autoref{tbMain}). But with joint end-to-end training enabled by soft
quantization, our \ETE:\vPIXOR consistently outperforms the separately trained PL++: \vPIXOR.

\begin{table}[!t]
\tabcolsep 3pt
\centering
\caption{\small \textbf{Ablation studies on the quantization-based pipeline with \vPIXOR.} We report \APBEV at IoU $=0.5$ / $0.7$ (in \%) of the \textbf{car} category. We divide our pipeline into two sub networks: Depth and Detector. $\surd$ means we set the sub network trainable and use its corresponding loss in join training. The best result per column is in {\color{blue} blue}. Best viewed in color.}
\vskip -5pt
\begin{tabular}{=c|+c|+c|+c|+c}
	Depth & Detector & Easy & Moderate & Hard \\ \hline
	\hline &  & 89.8 / 77.0  & 78.3 / 57.7& 69.5 / 53.8 \\ \hline
	$\surd$ &  & 89.9 / 76.9  & 78.7 / 58.0& 69.7 / 53.9 \\ \hline
	& $\surd$ &	90.2 / 78.1  & 79.2 /  58.9&  69.6 / 54.2 \\ \hline
	$\surd$ & $\surd$ & {\color{blue}94.6} / {\color{blue}80.4}  & {\color{blue}84.8} / {\color{blue}64.3} & {\color{blue}77.1} / {\color{blue}56.7} \\
    \hline
\end{tabular}
\label{tbAblation_PIXOR}
\end{table}

\begin{figure}[!t]
	\centering
	\small
	\centerline{\includegraphics[width=\linewidth]{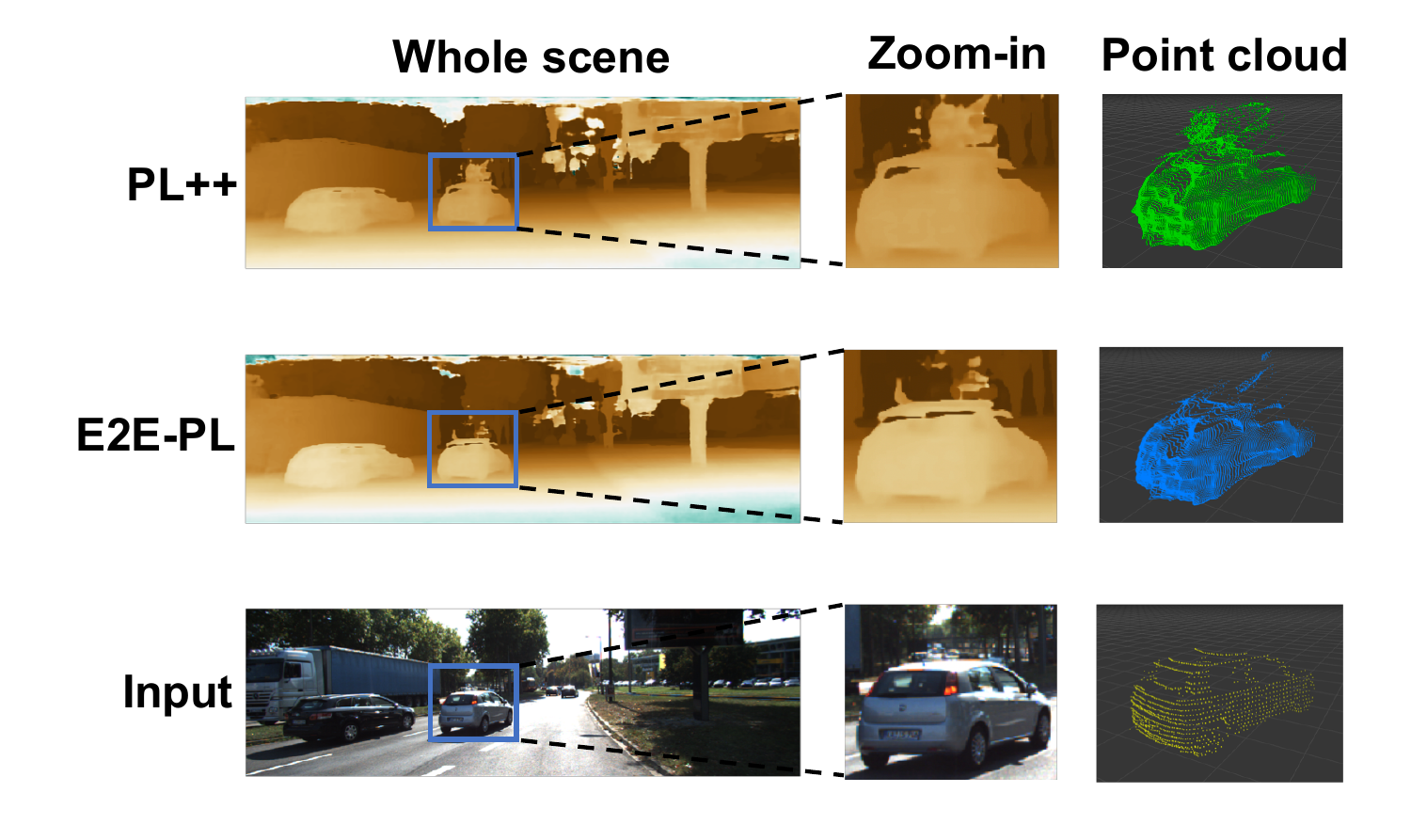}}
	\vskip-5pt
	\caption{\small \textbf{Qualitative results on depth estimation.} PL++ (image-only) has many misestimated pixels on the top of the car. By applying end-to-end training, the depth estimation around the cars is improved and the corresponding pseudo-LiDAR point cloud has much better quality. (Please zoom-in for the better view.)}
	\label{fig:depth}
	\vskip-10pt
\end{figure}

\begin{figure*}[!t]
    \centering
    \small
    \centerline{\includegraphics[width=0.95\textwidth]{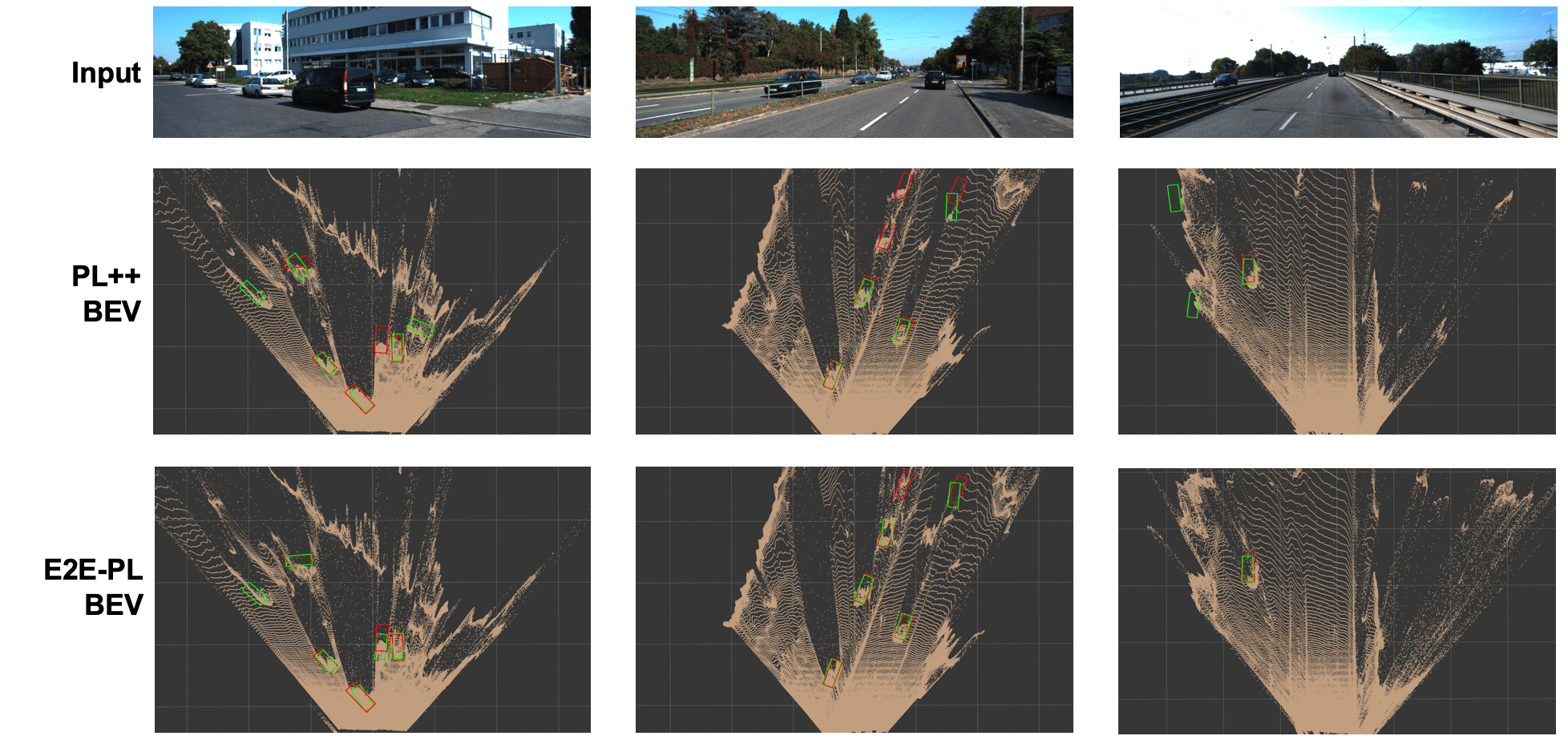}}
    \vskip-5pt
    \caption{\small\textbf{Qualitative results from the bird's-eye view.} The {\color{red}red} bounding boxes are the ground truth and the {\color{green}green} bounding boxes are the detection results. PL++ (image-only) misses many far-away cars and has poor bounding box localization. By applying end-to-end training, we get much accurate predictions (first and second columns) and reduce the false positive predictions (the third column).}
    \label{fig:detect} 
    \vskip-10pt
\end{figure*}

\subsection{Qualitative results}
\label{ssec:qual}
We show the qualitative results of the point-cloud-based and quantization-based pipelines.

\noindent\textbf{Depth visualization.}
We visualize the predicted depth maps and the corresponding point clouds converted from the depth maps by the quantization-based pipeline in \autoref{fig:depth}. For the original depth network shown in the first row, since the ground truth is very sparse, the depth prediction is not accurate and we observe clear misestimation (artifact) on the top of the car: there are massive misestimated 3D points on the top of the car. 
By applying end-to-end joint training, the detection loss on cars enforces the depth network to reduce the misestimation and guides it to generate more accurate point clouds. As shown in the second row, both the depth prediction quality and the point cloud quality are greatly improved. The last row is the input image to the depth network, the zoomed-in patch of a specific car, and its corresponding LiDAR ground truth point cloud.

\noindent\textbf{Detection visualization.} We also show the qualitative comparisons of the detection results, as illustrated in \autoref{fig:detect}. From the BEV, the ground truth bounding boxes are marked in red and the predictions are marked in green. In the first example (column), 
pseudo-LiDAR++ (PL++) misses one car in the middle, and gives poor localization of the far-away cars. Our \ETE could detect all the cars, and gives accurate predictions on the farthest car. For the second example, the result is consistent for the far-away cars where \ETE detects more cars and localize them more accurately. Even for the nearby cars, \ETE gives better results.
The third example indicates a case where there is only one ground truth car. Our \ETE does not have any false positive predictions.

\subsection{Other results}
\label{ssec:other}
\noindent\textbf{Speed.} The inference time of our method is similar to pseudo-LiDAR and is determined by the stereo and detection networks. The soft quantization module (with 26 neighboring bins) only computes the RBF weights between a point and 27 bins (roughly $N=300,000$ points per scene), followed by grouping the weights of points into bins. The complexity is $O(N)$, and both steps can be parallelized. Using a single GPU with PyTorch implementation, \ETE: \PRCNN takes 0.49s/frame and \ETE: \vPIXOR takes 0.55s/frame, within which \SDN (stereo network) takes 0.39s/frame and deserves further study to speed it up (e.g., by code optimization, network pruning, etc).
\noindent\textbf{Others.} See the Supplementary Material for more details and results.

%% file: disc.tex
\section{Conclusion and Discussion}
\label{sec:disc}
In this paper, we introduced an end-to-end training framework for pseudo-LiDAR~\cite{pseudoLiDAR,you2019pseudo}. Our proposed framework can work for 3D object detectors taking either  the direct point cloud inputs or quantized structured inputs. The resulting models set a new state of the art in image based 3D object detection and further narrow the remaining accuracy gap between stereo and LiDAR based sensors. Although it will probably always be beneficial to include active sensors like LiDARs in addition to passive cameras~\cite{you2019pseudo}, it seems possible that the benefit may soon be too small to justify large expenses. Considering the KITTI benchmark, it is worth noting that the stereo pictures are of relatively low resolution and only few images contain (labeled) far away objects.  It is quite plausible that higher resolution images with a higher ratio of far away cars would result in further detection improvements, especially in the hard (far away and heavily occluded) category. 

%% file: supplementary.tex
\renewcommand{\thesection}{S\arabic{section}}
\renewcommand{\thetable}{S\arabic{table}}
\renewcommand{\thefigure}{S\arabic{figure}}
\definecolor{lightpurple}{RGB}{178,145,226}
\definecolor{lightblue}{RGB}{0,128,255}
\definecolor{yellow}{rgb}{1,1,0}

\def\WIDTHFIVE {0.19\textwidth}
\def\WIDTHTHREE {0.315\textwidth}
\def\WIDTHTWO {0.45\textwidth}

We provide details and results omitted in the main text.

\begin{itemize}
	\item \autoref{supp-sec: pedestrian}: results in pedestrians and cyclists (\autoref{ssec:results} of the main paper).
	\item \autoref{supp-sec:range}: results at different depth ranges (\autoref{ssec:results} of the main paper).
	\item \autoref{supp-sec:test}: KITTI testset results (\autoref{ssec:results} of the main paper).
	\item \autoref{supp-qual}: additional qualitative results (\autoref{ssec:qual} of the main paper).
	\item \autoref{suppl-sec:grad}: gradient visualization (\autoref{ssec:qual} of the main paper).
	\item \autoref{suppl-sec:other}: other results (\autoref{ssec:other} of the main paper).
\end{itemize}


\section{Results on Pedestrians and Cyclists}
\label{supp-sec: pedestrian}
In addition to 3D object detection on Car category, in \autoref{tb::ped_cyc} we show the results on Pedestrian and Cyclist categories in KITTI object detection validation set~\cite{geiger2013vision,geiger2012we}. To be consistent with the main paper, we apply \PRCNN~\cite{shi2019pointrcnn} as the object detector. Our approach (\ETE) outperforms the baseline one without end-to-end training (PL++)~\cite{you2019pseudo} by a notable margin for image-based 3D detection.

\begin{table}[htbp]
\centering
\small
\tabcolsep 3.5pt
\captionsetup{font=small}
\begin{tabular}{c|c|c|c|c}
Category                    & Model  & Easy      & Moderate  & Hard      \\ \hline
\multirow{2}{*}{Pedestrian} & PL++   & 31.9 / 26.5 & 25.2 / 21.3 & 21.0 / 18.1 \\ 
                            & \ETE  & {\color{blue}\textbf{35.7}} / {\color{blue}\textbf{32.3}} & {\color{blue}\textbf{27.8}} / {\color{blue}\textbf{24.9}} & {\color{blue}\textbf{23.4}} / {\color{blue}\textbf{21.5}} \\ \hline
\multirow{2}{*}{Cyclist}    & PL++   & 36.7 / 33.5 & 23.9 / 22.5 & 22.7 / 20.8 \\ 
                            & \ETE  & {\color{blue}\textbf{42.8}} / {\color{blue}\textbf{38.4}} & {\color{blue}\textbf{26.2}} / {\color{blue}\textbf{24.1}} & {\color{blue}\textbf{24.5}} / {\color{blue}\textbf{22.7}} \\ \hline
\end{tabular}
\caption{\textbf{Results on pedestrians and cyclists (KITTI validation set).} We report \APBEV ~/ \AP (in \%) of the two categories at IoU=0.5, following existing works~\cite{shi2019pointrcnn,pseudoLiDAR}. PL++ denotes the \PL++ pipeline with images only (i.e., SDN alone)~\cite{pseudoLiDAR}. Both approaches use \PRCNN~\cite{shi2019pointrcnn} as the object detector. \label{tb::ped_cyc}}
\end{table}


\section{Evaluation at Different Depth Ranges}
\label{supp-sec:range}
We analyze 3D object detection of Car category for ground truths at different depth ranges (i.e., 0-30 or 30-70 meters). We report results with the point-cloud-based pipelines in \autoref{tb::prcnn_range} and the quantization-based pipeline in \autoref{tb::pixor_range}. \ETE achieves better performance at both depth ranges (except for 30-70 meters, moderate, \AP). Specifically, on \APBEV, the relative gain between \ETE and the baseline becomes larger for the far-away range and the hard setting.

\begin{table}[htbp]
\centering
\small
\tabcolsep 3pt
\captionsetup{font=small}
\begin{tabular}{c|c|c|c|c||c}
Range             &Model  & Easy      & Moderate  & Hard      & \#Objs                \\ \hline
\multirow{2}{*}{0-30} & PL++ & 82.9 / 68.7 & 76.8 / 64.1 & 67.9 / 55.7 & \multirow{2}{*}{7379} \\
                       & \ETE & {\color{blue}\textbf{86.2}} / {\color{blue}\textbf{72.7}} & {\color{blue}\textbf{78.6}} / {\color{blue}\textbf{66.5}} & {\color{blue}\textbf{69.4}} / {\color{blue}\textbf{57.7}} &                       \\ \hline
\multirow{2}{*}{30-70}& PL++ & 19.7 / 11.0 & 29.5 / 18.1 & 27.5 / 16.4 & \multirow{2}{*}{3583} \\ 
                       & \ETE & {\color{blue}\textbf{23.8}} / {\color{blue}\textbf{15.1}} & {\color{blue}\textbf{31.8}} / {\color{blue}\textbf{18.0}} & {\color{blue}\textbf{31.0}} / {\color{blue}\textbf{16.9}} &                       \\ \hline  
\end{tabular}
\caption{\textbf{3D object  detection via the point-cloud-based pipeline with \PRCNN at different depth ranges.} We report \APBEV ~/ \AP (in \%) of the \textbf{car} category at IoU=0.7, using \PRCNN for detection. In the last column we show the number of car objects in KITTI object validation set within different ranges. \label{tb::prcnn_range}}
\end{table}

\begin{table}[htbp]
\centering
\small
\tabcolsep 3pt
\captionsetup{font=small}
\begin{tabular}{c|c|c|c|c||c}
Range             &Model  & Easy      & Moderate  & Hard      & \#Objs                \\ \hline
\multirow{2}{*}{0-30} & PL++ & 81.4 / -  & 75.5 / - & 65.8 / - & \multirow{2}{*}{7379} \\
                       & \ETE  & {\color{blue}\textbf{82.1} / -} & {\color{blue}\textbf{76.4} / -} & {\color{blue}\textbf{67.5} / -} &                       \\ \hline
\multirow{2}{*}{30-70}& PL++ & 26.1 / - & 23.9 / - & 20.5 / - & \multirow{2}{*}{3583} \\ 
                       & \ETE  & {\color{blue}\textbf{26.8} / -} & {\color{blue}\textbf{36.1} / -} & {\color{blue}\textbf{31.7} / -} &                       \\ \hline        
\end{tabular}
\caption{\textbf{3D object detection via the quantization-based pipeline with \vPIXOR at different depth ranges.} The setup is the same as in \ref{tb::prcnn_range}, except that \vPIXOR does not have height prediction and therefore no \AP is reported. \label{tb::pixor_range}}
\end{table}

\begin{figure*}[htbp]
	\begin{center}
		{  
			\subfigure[PL++ on 3D Track (\AP)]
			{           
				\includegraphics[width=0.48\textwidth]{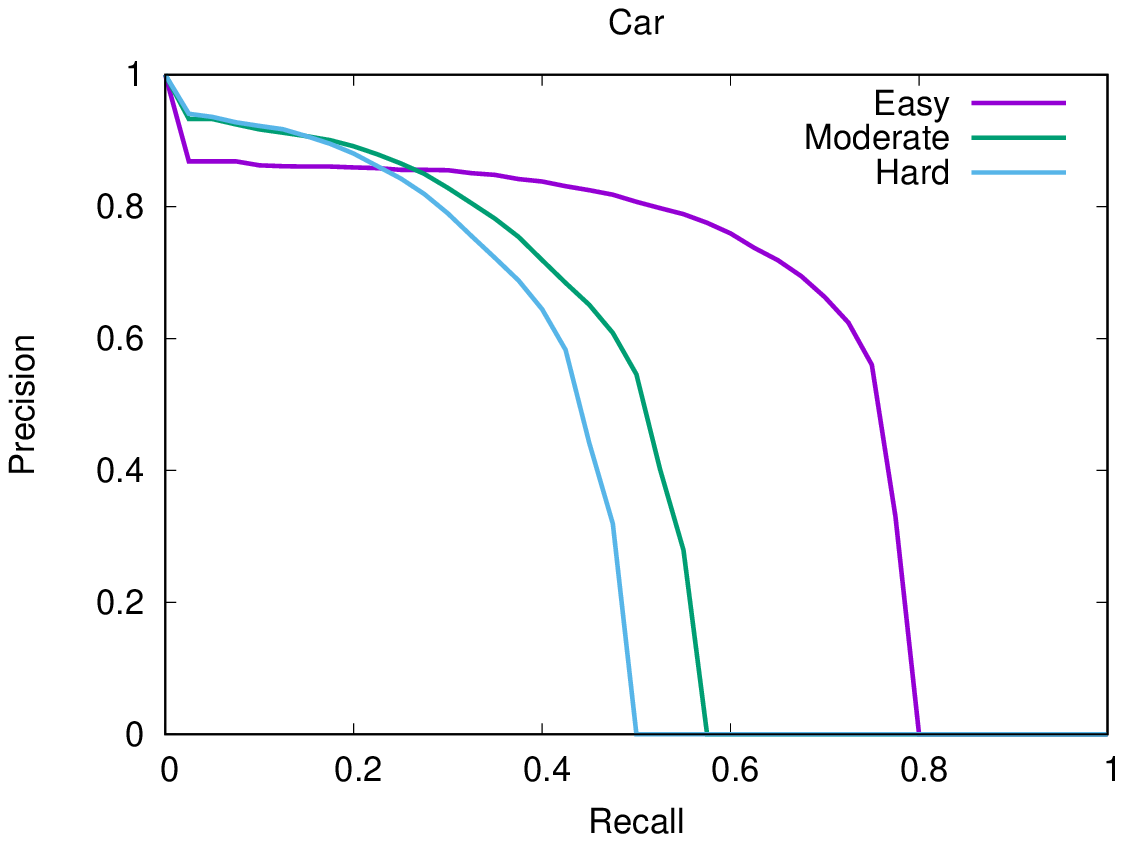}
			}
			\subfigure[E2E-PL on 3D Track (\AP)]
			{
				\includegraphics[width=0.48\textwidth]{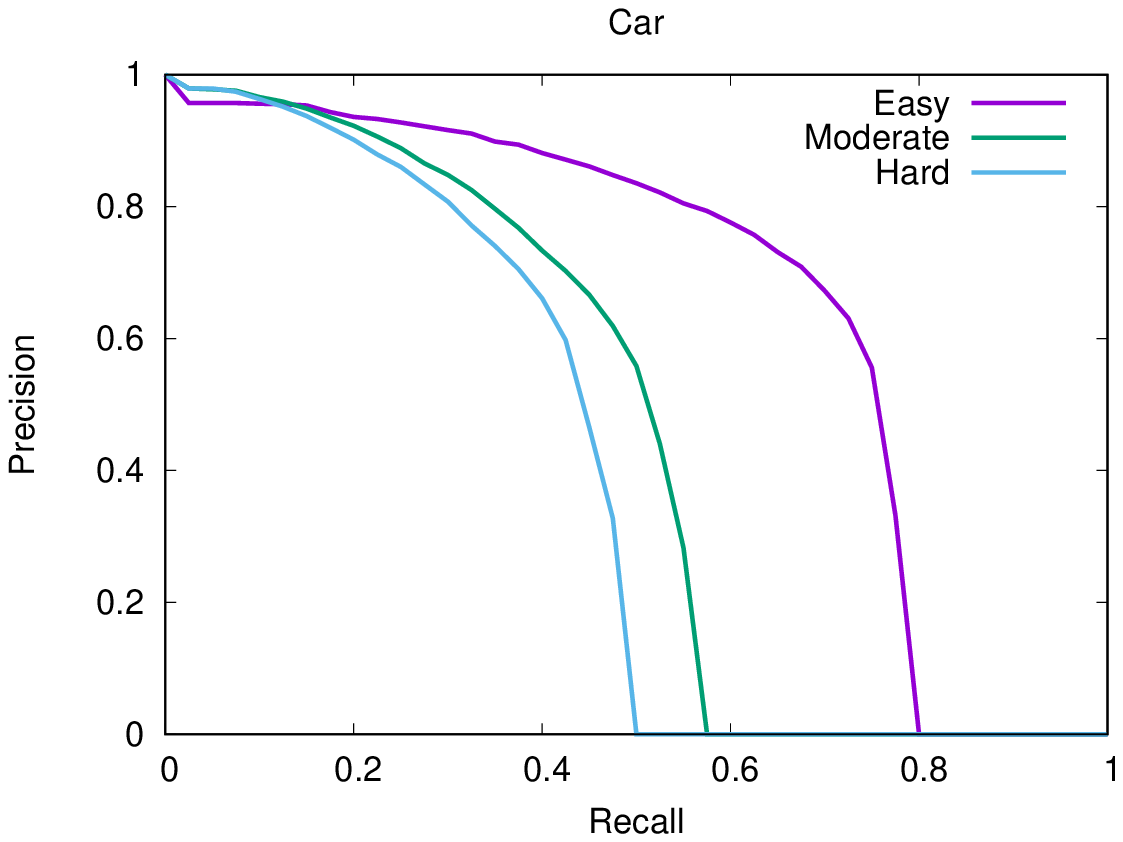}
			}       
			\\
			\subfigure[PL++ on BEV Track (\APBEV)]
			{          
				\includegraphics[width=0.48\textwidth]{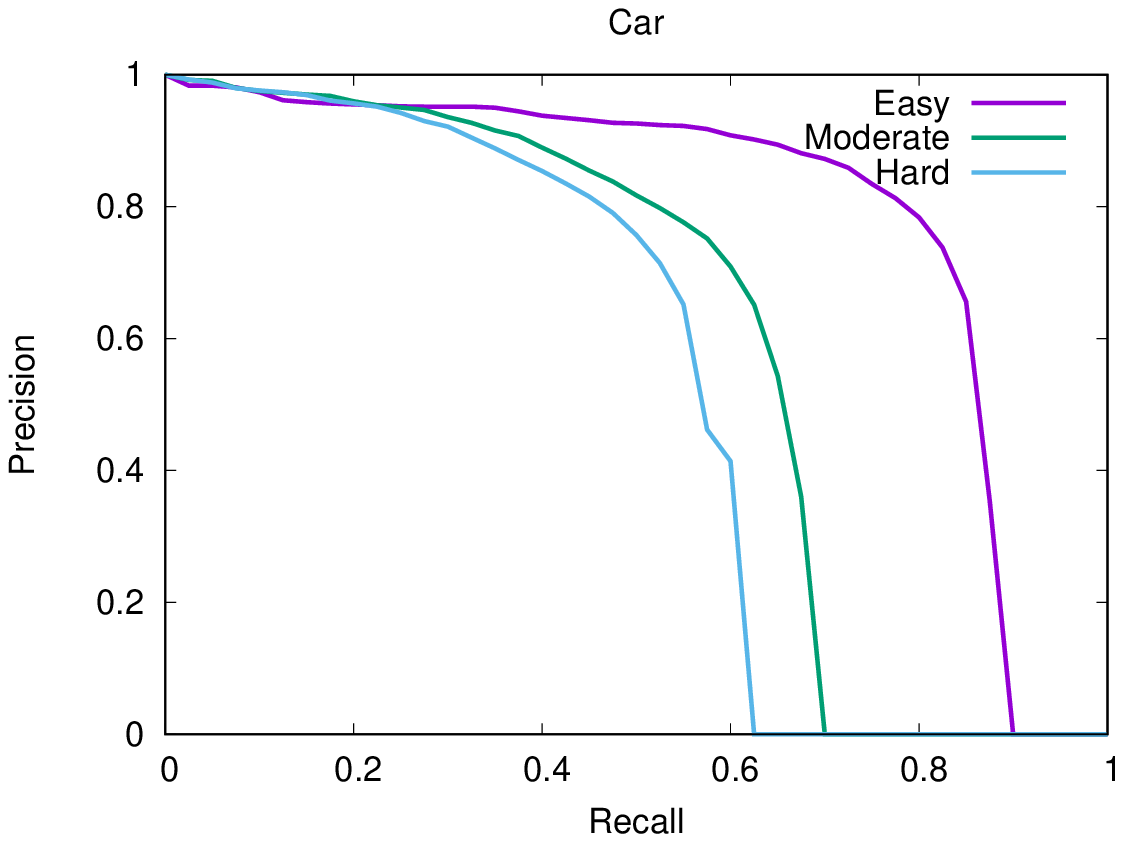}
			}    
			\subfigure[E2E-PL on BEV Track (\APBEV)]
			{          
				\includegraphics[width=0.48\textwidth]{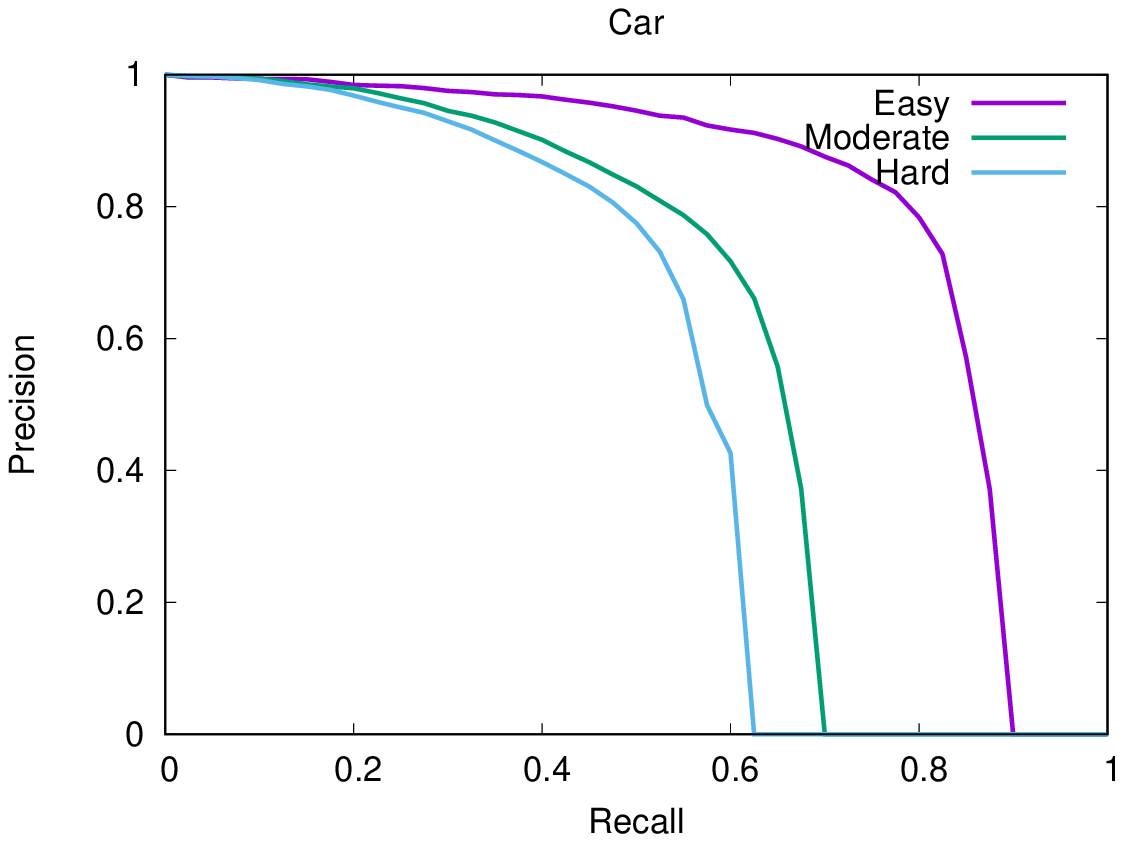}
			}                       
		} 
	\end{center}
	\caption{\textbf{Precision-recall curves on KITTI test dataset.} We here compare E2E-PL with PL++ on the 3D object detection track and bird's eye view detection track.}
	\label{fig:cmp1}
\end{figure*}

\section{On KITTI Test Set}
\label{supp-sec:test}

In \autoref{fig:cmp1}, we compare the precision-recall curves of our \ETE and \PL++ (named Pseudo-LiDAR V2 on the leaderboard). On the 3D object detection track (first row of \autoref{fig:cmp1}), \PL++ has a notable drop of precision on easy cars even at low recalls, meaning that \PL++ has many high-confident false positive predictions. The same situation happens to moderate and hard cars. Our E2E-PL suppresses the false positive predictions, resulting in more smoother precision-recall curves. On the bird's-eye view detection track (second row of \autoref{fig:cmp1}), the precision of \ETE is over 97\% within recall interval 0.0 to 0.2, which is higher than the precision of \PL++, indicating that \ETE has fewer false positives.


\section{Additional Qualitative Results}
\label{supp-qual}
\begin{figure*}[htbp]
   \begin{center}
      {  
         \subfigure
         {           
            \includegraphics[width=0.32\textwidth]{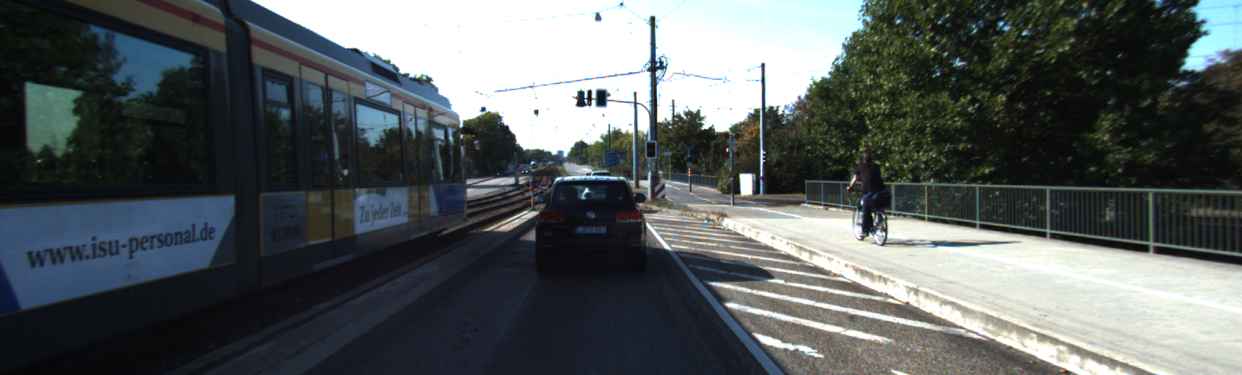}
         }
         \subfigure
         {          
            \includegraphics[width=0.32\textwidth]{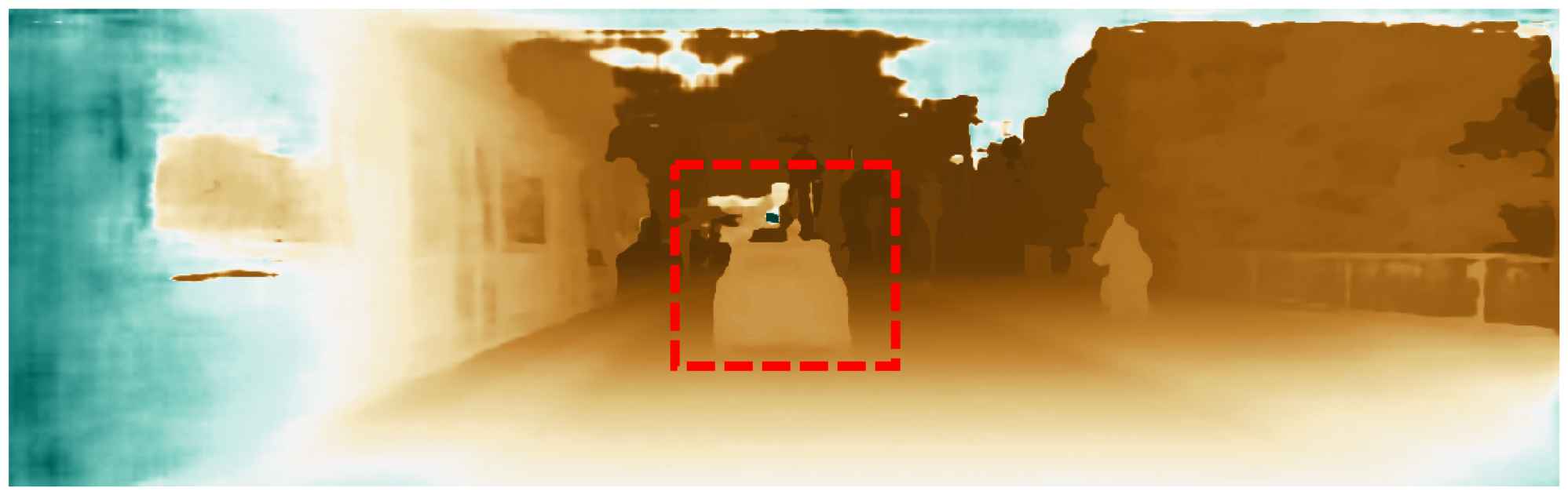}
         }    
         \subfigure
         {          
            \includegraphics[width=0.32\textwidth]{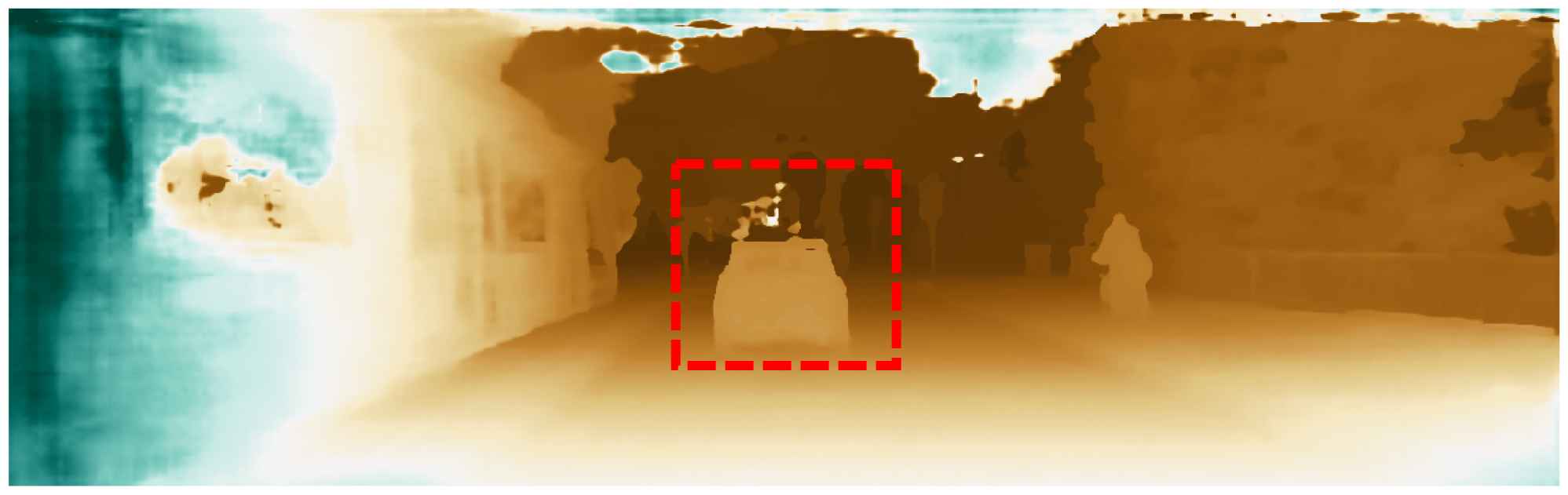}
         }\\ \vspace{-3mm}
         
         \subfigure
         {           
            \includegraphics[width=0.32\textwidth]{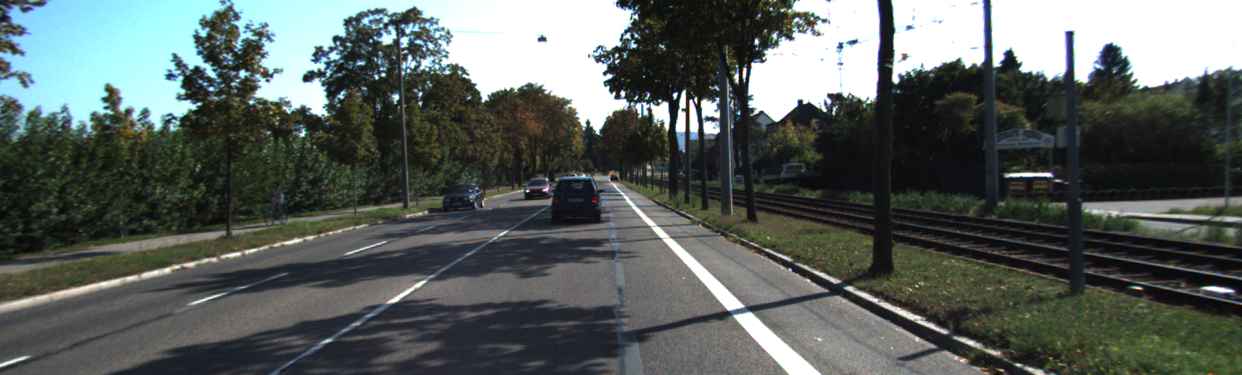}
         }
         \subfigure
         {          
            \includegraphics[width=0.32\textwidth]{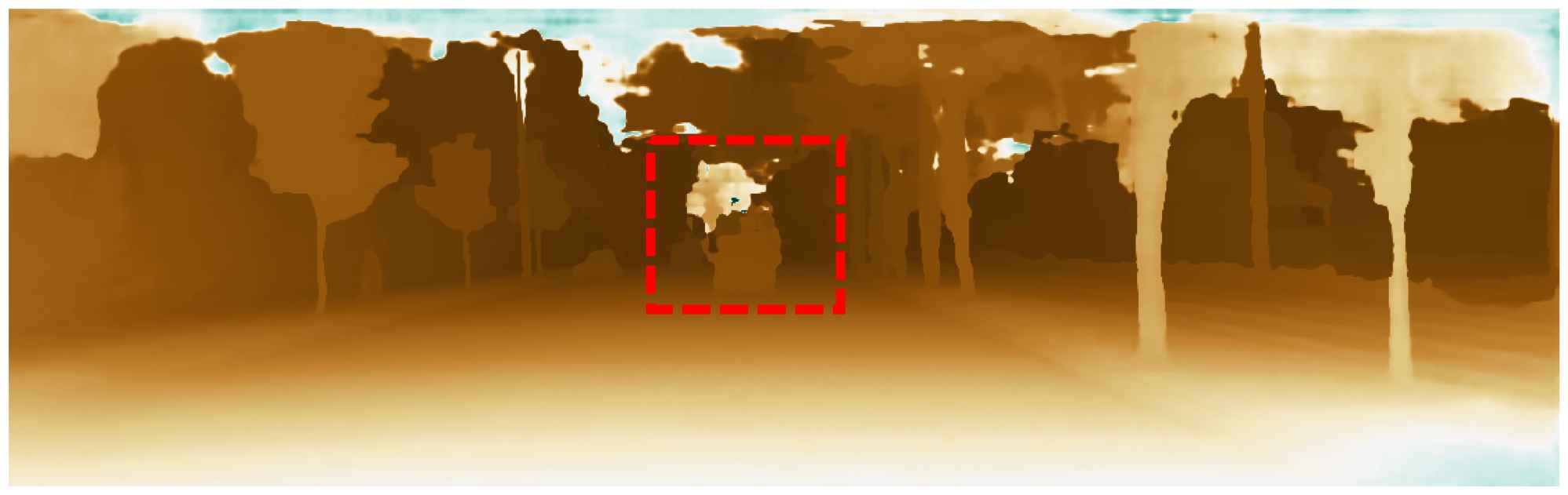}
         }    
         \subfigure
         {          
            \includegraphics[width=0.32\textwidth]{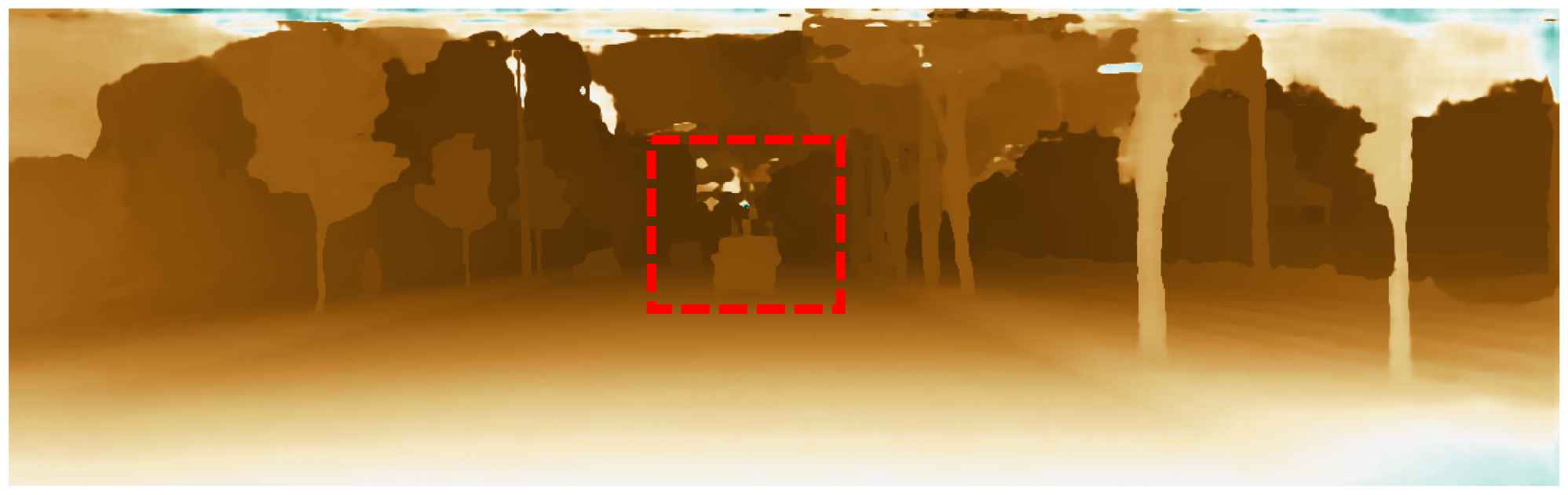}
         }\\ \vspace{-3mm}
         
         \subfigure
         {           
            \includegraphics[width=0.32\textwidth]{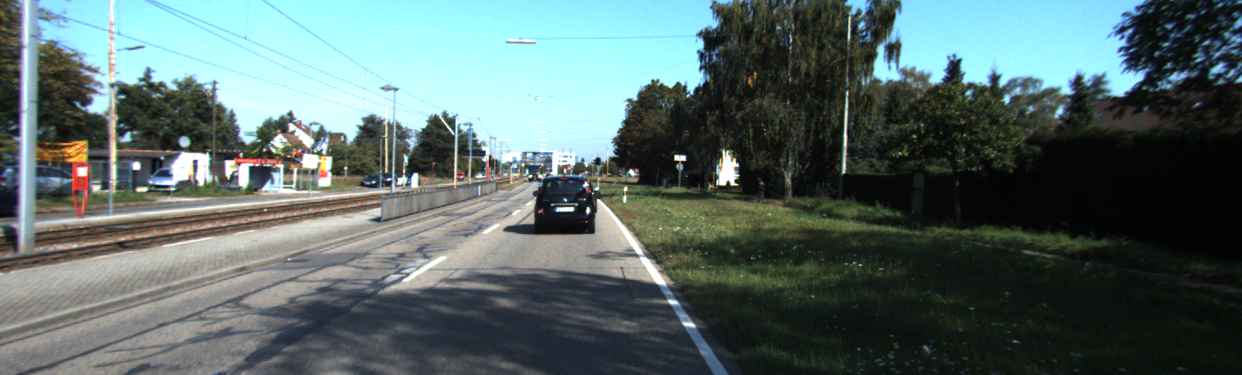}
         }
         \subfigure
         {          
            \includegraphics[width=0.32\textwidth]{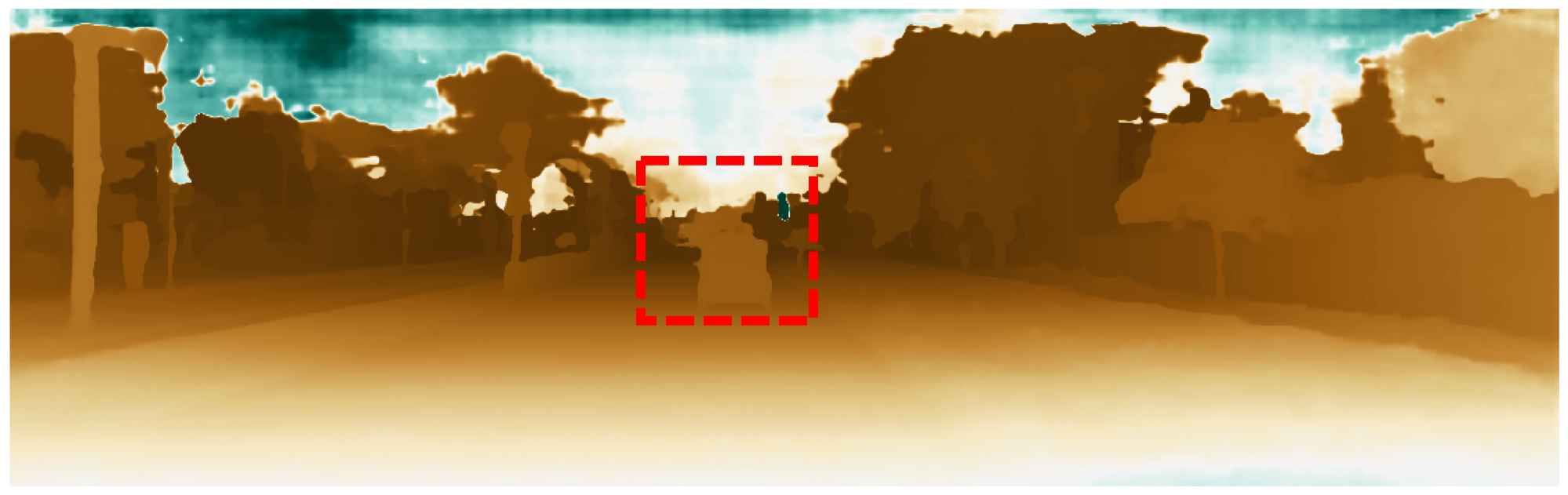}
         }    
         \subfigure
         {          
            \includegraphics[width=0.32\textwidth]{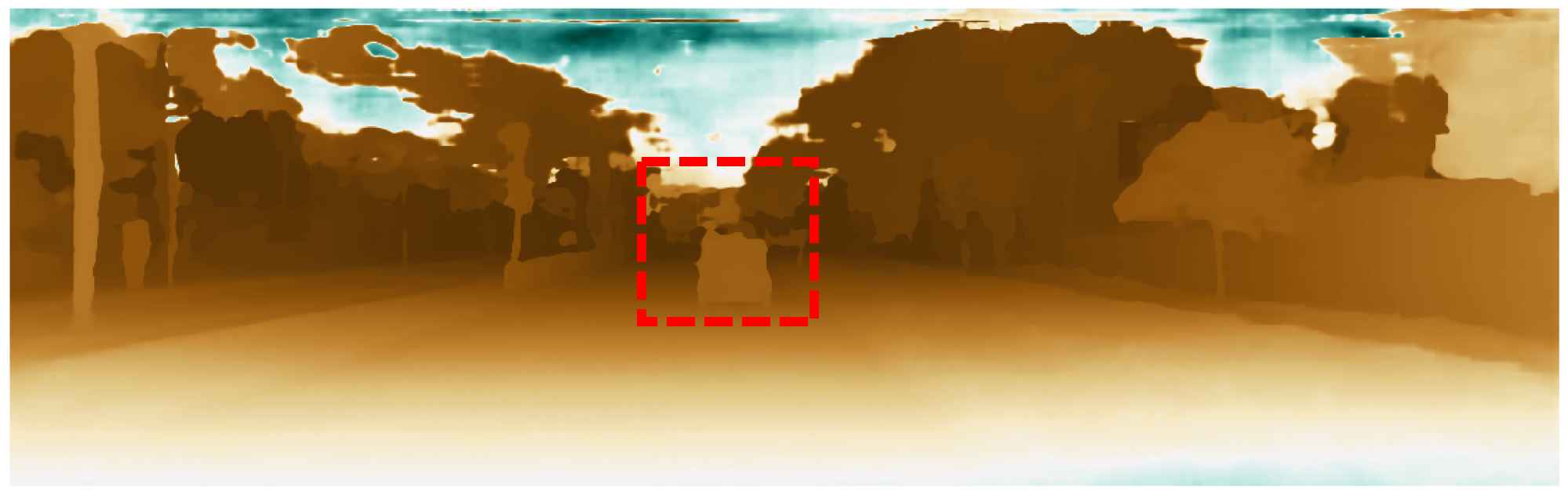}
         }\\ \vspace{-3mm}
         
         \subfigure
         {           
            \includegraphics[width=0.32\textwidth]{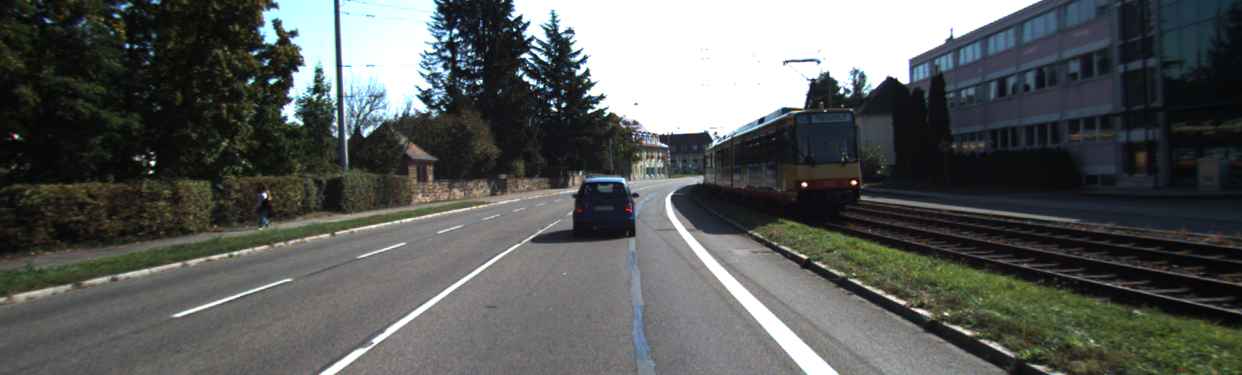}
         }
         \subfigure
         {          
            \includegraphics[width=0.32\textwidth]{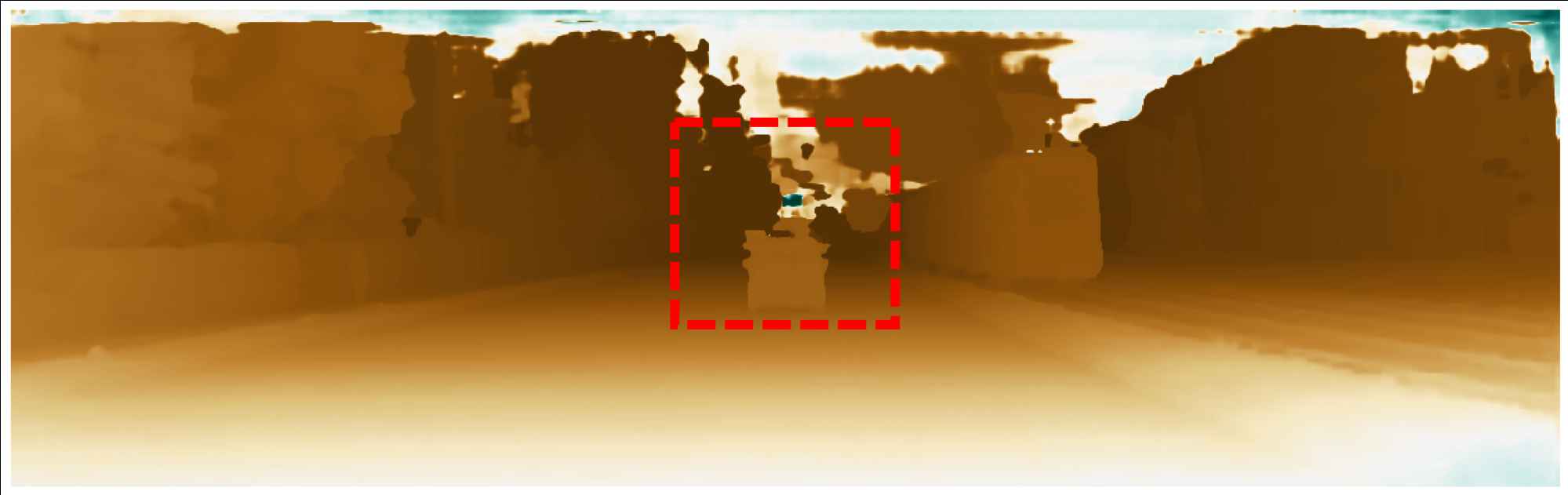}
         }    
         \subfigure
         {          
            \includegraphics[width=0.32\textwidth]{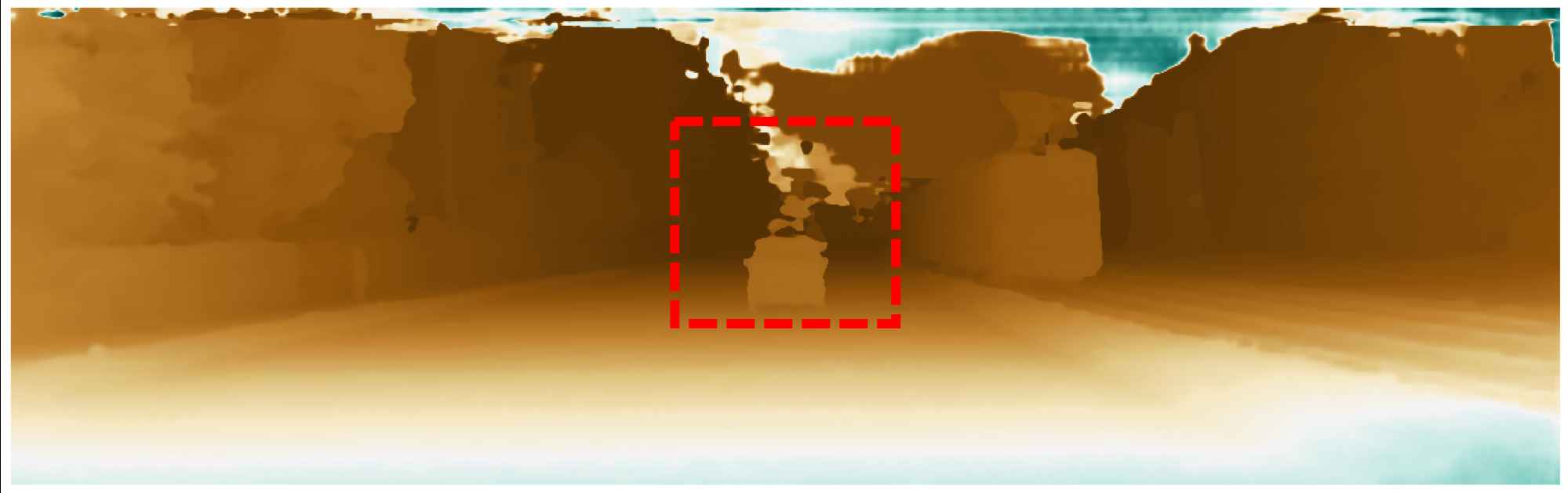}
         }\\ \vspace{-3mm}
         \addtocounter{subfigure}{-12}
         \subfigure[Input]
         {           
            \includegraphics[width=0.32\textwidth]{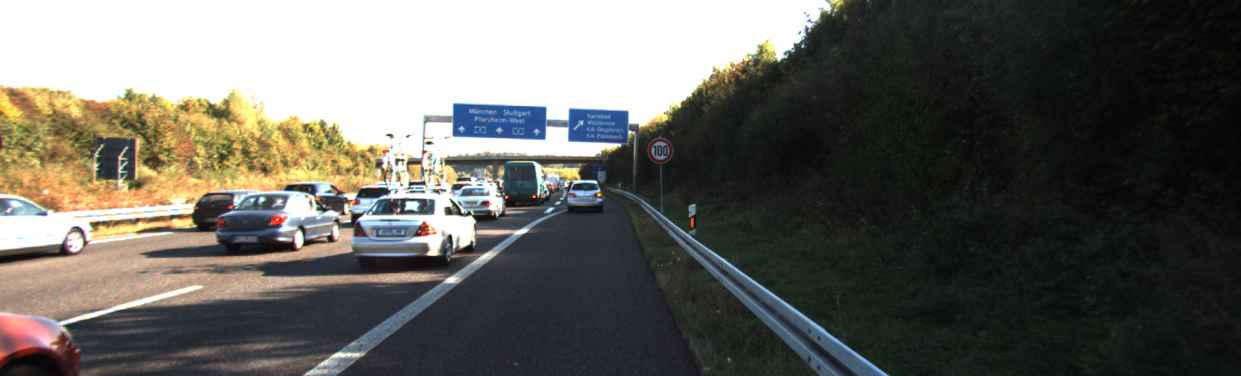}
         }
         \subfigure[Depth estimation of PL++]
         {          
            \includegraphics[width=0.32\textwidth]{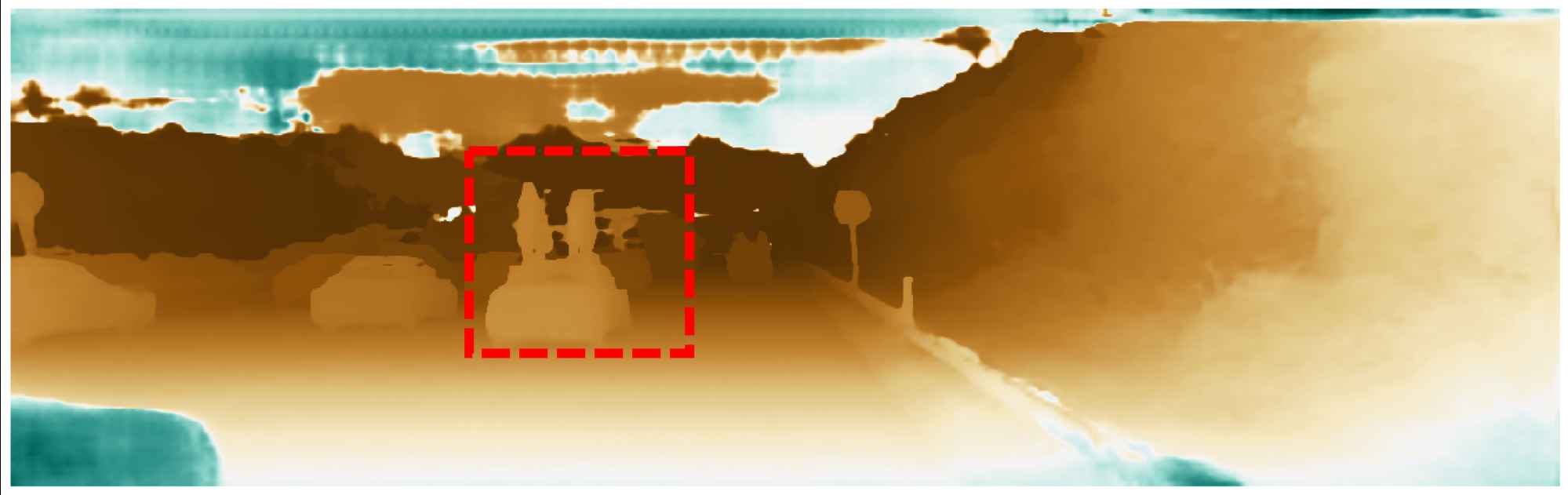}
         }    
         \subfigure[Depth estimation of E2E-PL]
         {          
            \includegraphics[width=0.32\textwidth]{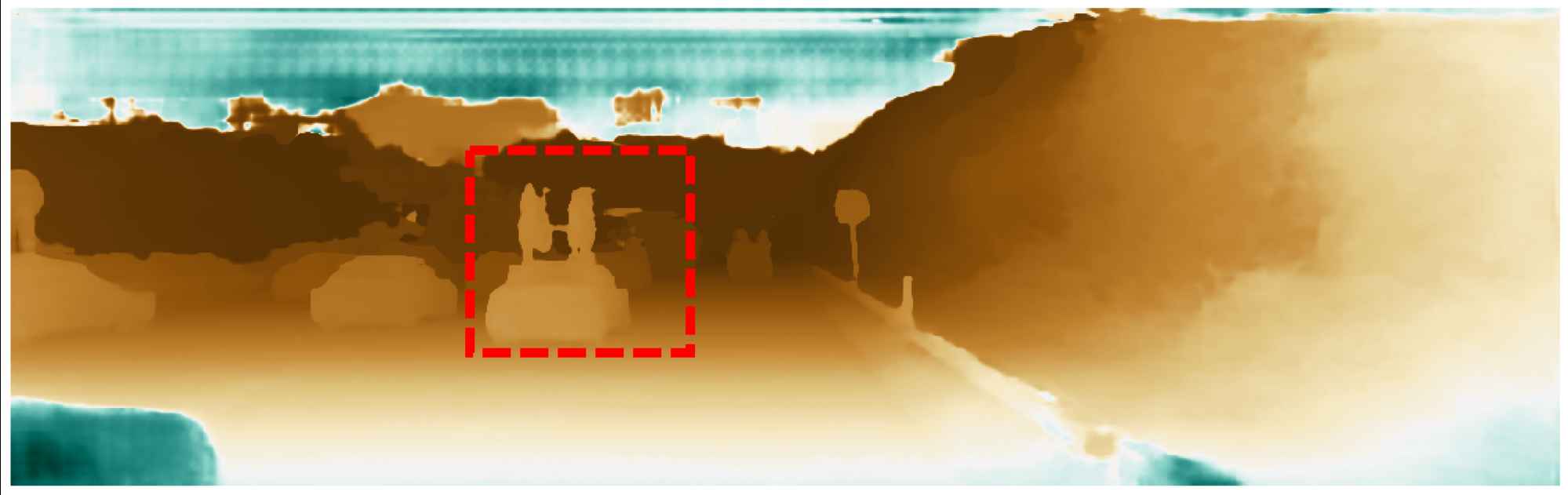}
         }
      } 
   \end{center}
   \vspace{-5mm}
   \caption{\textbf{Qualitative comparison of depth estimation.} We here compare PL++ with E2E-PL.}
   \label{fig:cmp2}
\end{figure*}

\begin{figure*}[htbp]
   \begin{center}
      {  
         \subfigure[Image]
         {           
            \includegraphics[width=0.32\textwidth]{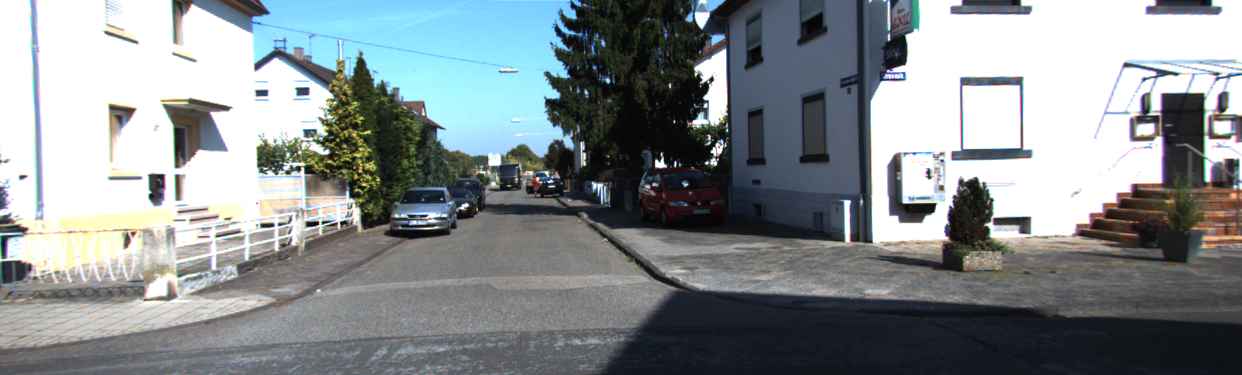}
         }
         \subfigure[Image]
         {          
            \includegraphics[width=0.32\textwidth]{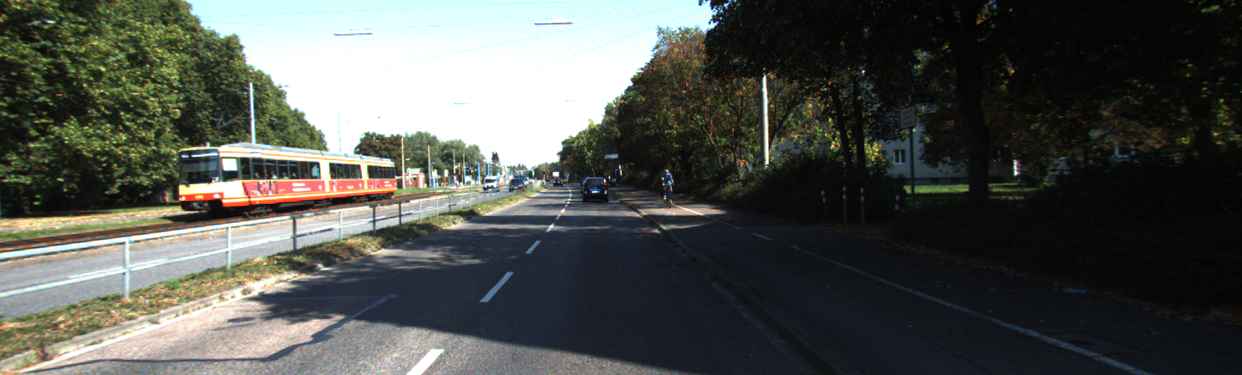}
         }    
         \subfigure[Image]
         {          
            \includegraphics[width=0.32\textwidth]{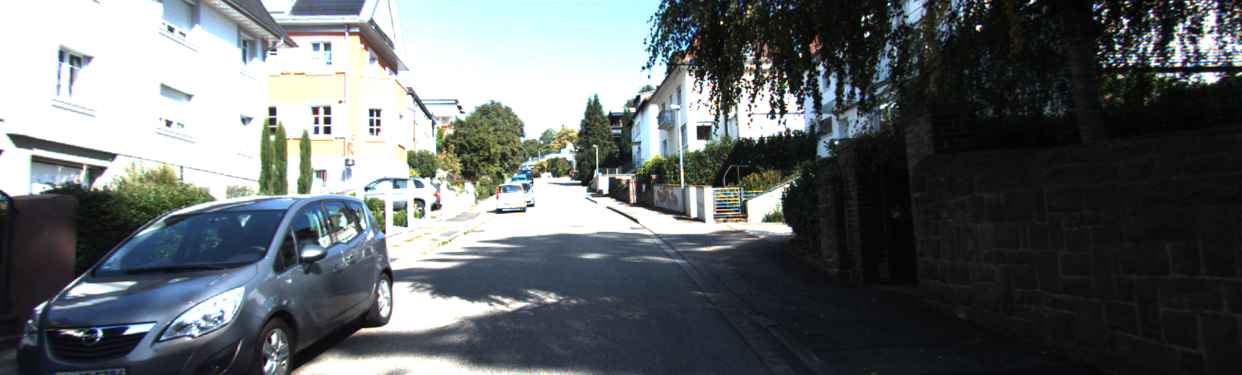}
         }\\ \vspace{-3mm}
         \subfigure[PL++ Detection]
         {           
            \includegraphics[width=0.32\textwidth]{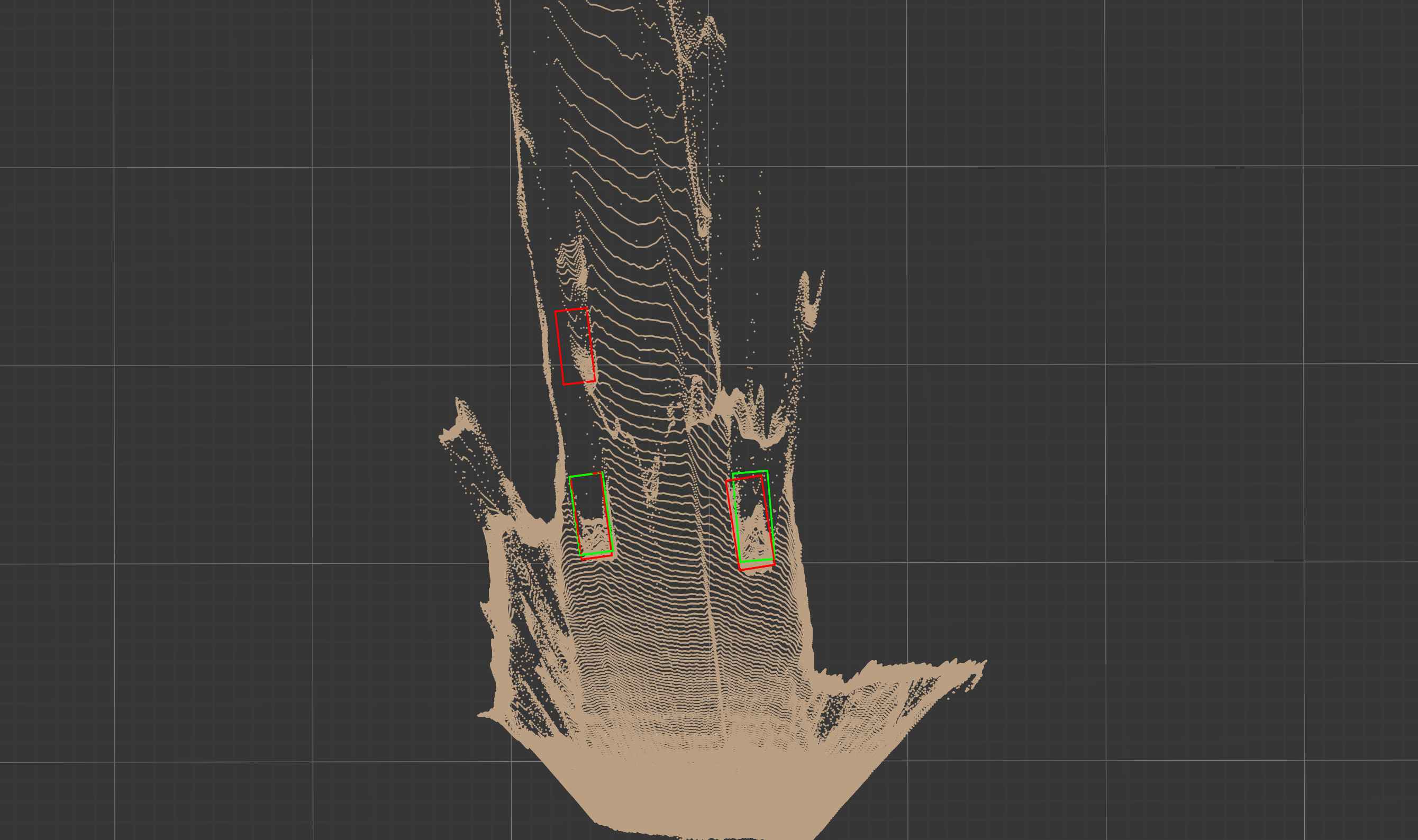}
         }
         \subfigure[PL++ Detection]
         {          
            \includegraphics[width=0.32\textwidth]{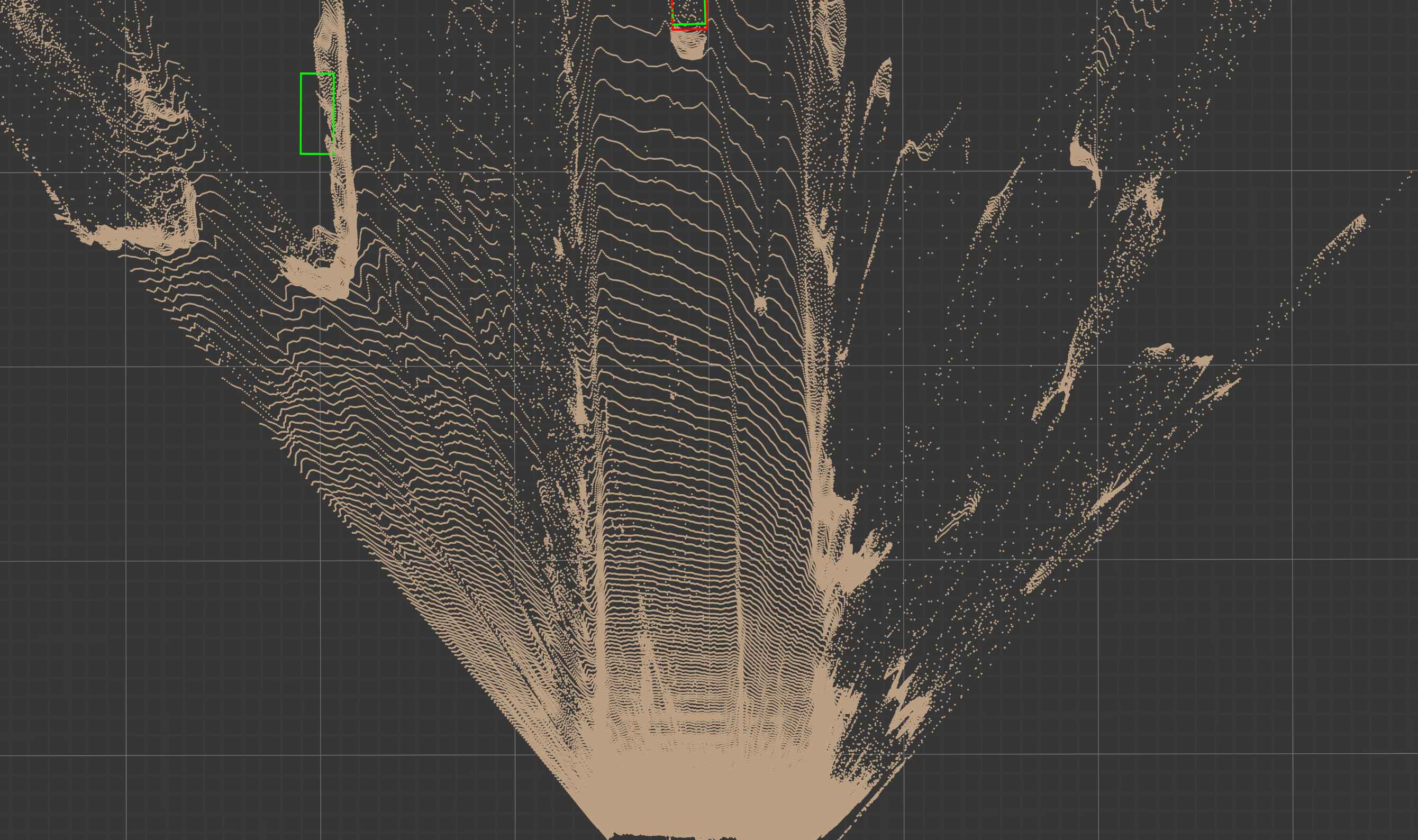}
         }    
         \subfigure[PL++ Detection]
         {          
            \includegraphics[width=0.32\textwidth]{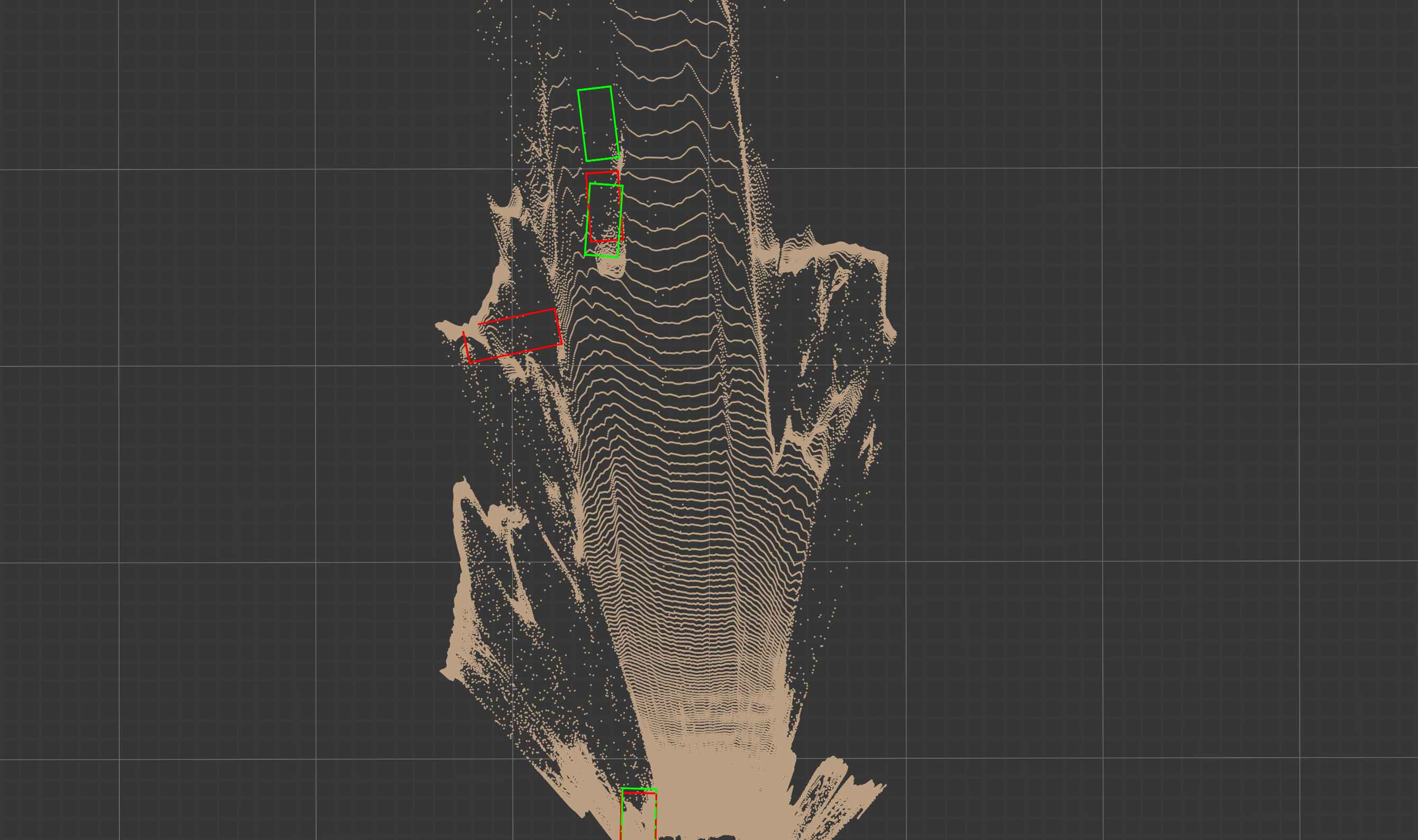}
         }\\ \vspace{-3mm}
         \subfigure[E2E-PL Detection]
         {           
            \includegraphics[width=0.32\textwidth]{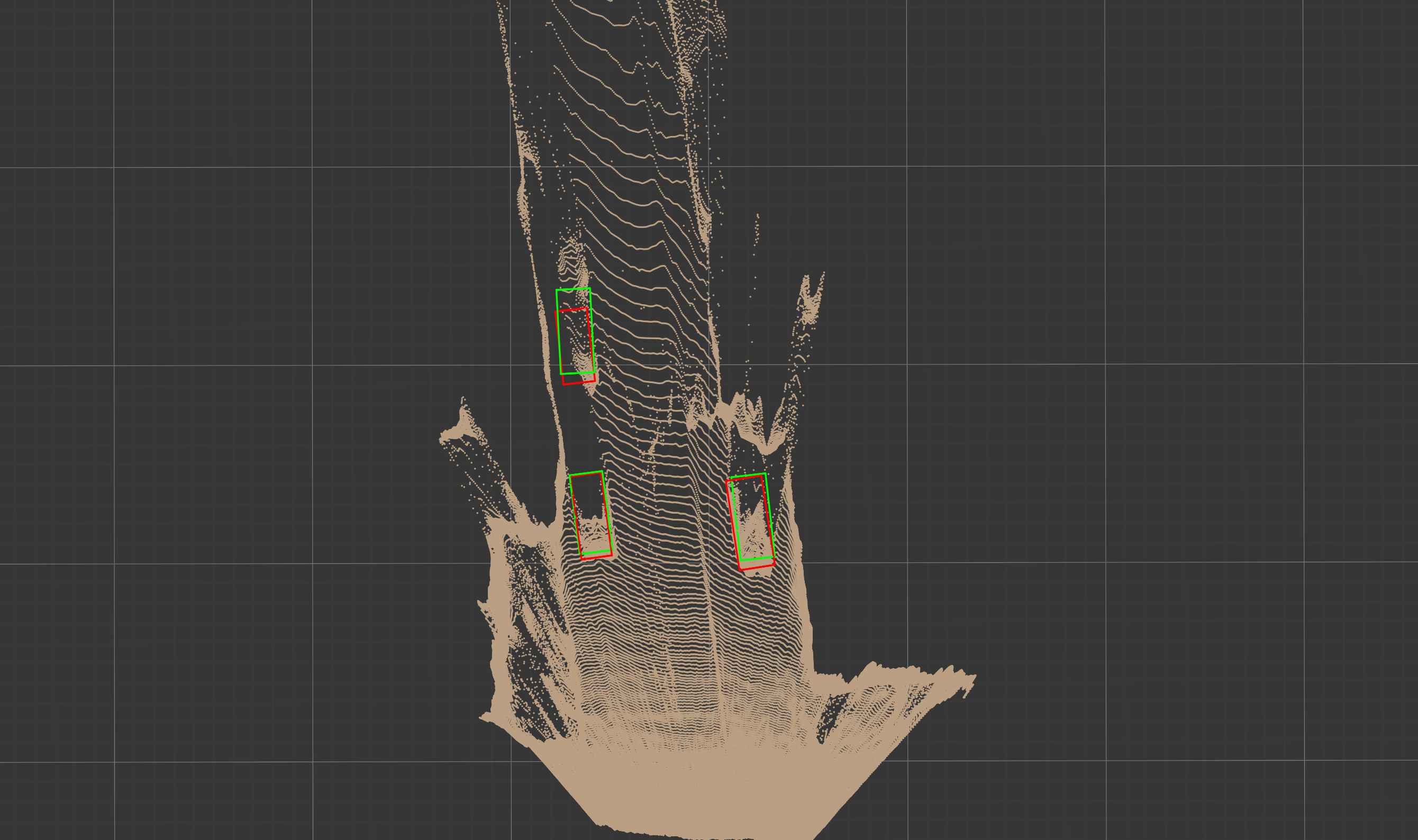}
         }
         \subfigure[E2E-PL Detection]
         {          
            \includegraphics[width=0.32\textwidth]{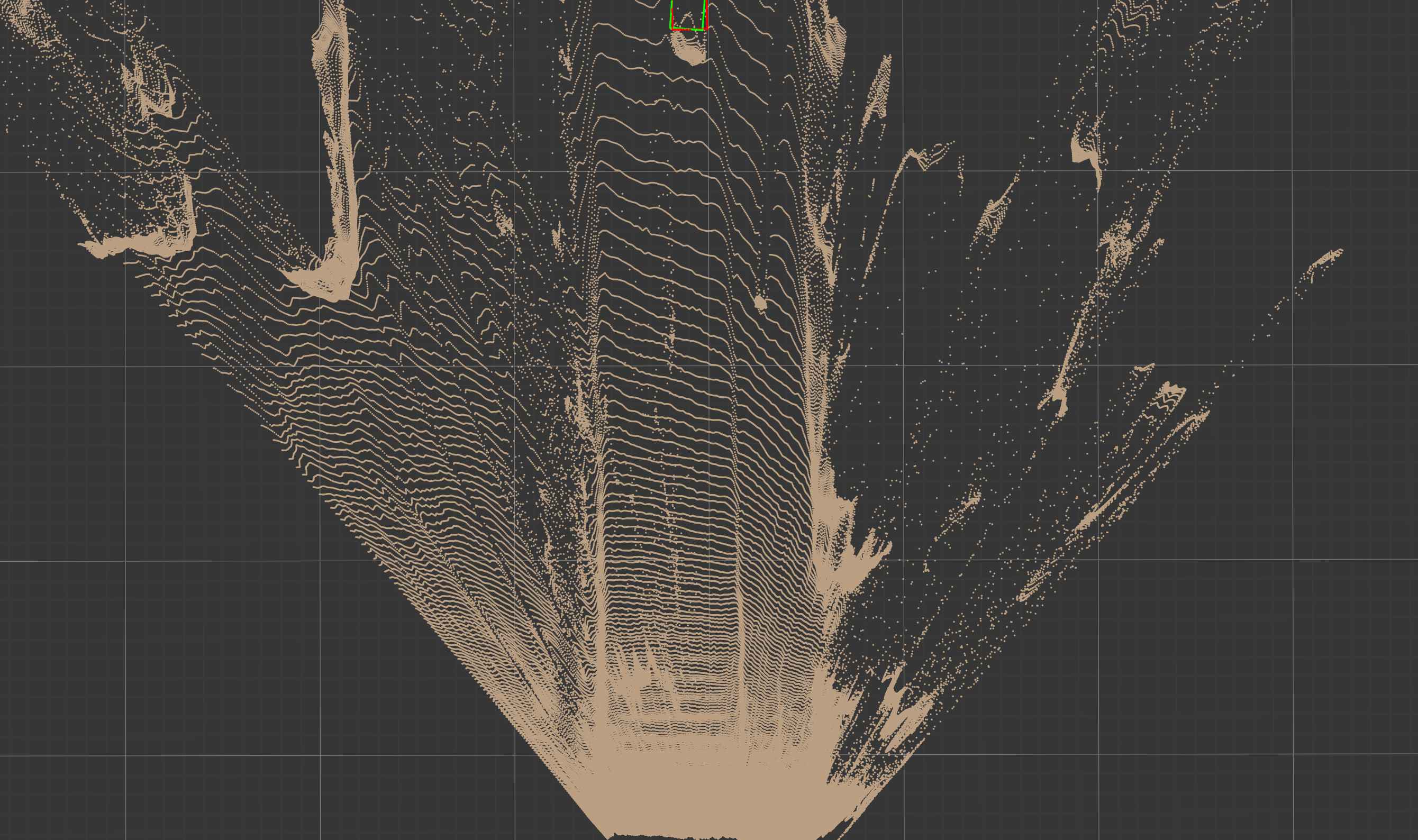}
         }    
         \subfigure[E2E-PL Detection]
         {          
            \includegraphics[width=0.32\textwidth]{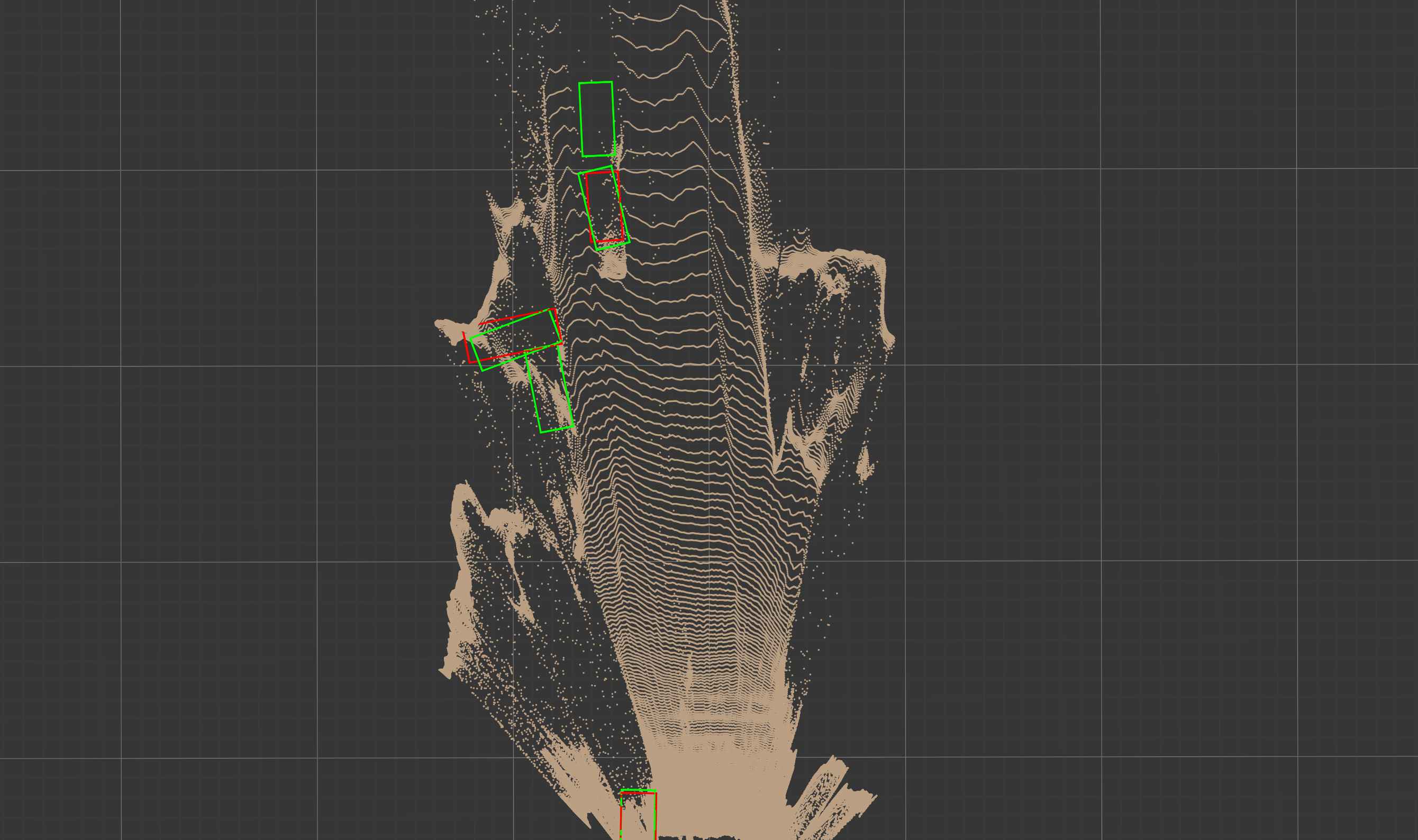}
         } \vspace{-3mm}
      } 
   \end{center}
   \caption{\textbf{Qualitative comparison of detection results.} We here compare PL++ with E2E-PL. The red bounding boxes are ground truth and the green bounding boxes are predictions.}
   \label{fig:cmp3}
\end{figure*}

We show more qualitative depth comparisons in \autoref{fig:cmp2}. We use red bounding boxes to highlight the depth improvement in car related areas. We also show detection comparisons in \autoref{fig:cmp3}, where our \ETE has fewer false positive and negative predictions.
\vspace{-5mm}

\vspace{3mm}
\section{Gradient Visualization on Depth Maps}
\label{suppl-sec:grad}
\begin{figure*}[htbp]
   \begin{center}
      {  
         \subfigure
         {           
            \includegraphics[width=0.48\textwidth]{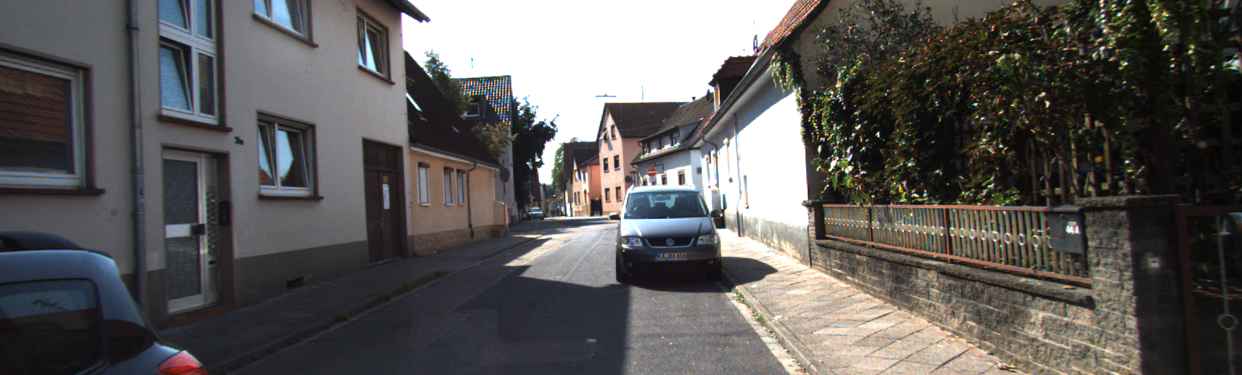}
         }
         \subfigure
         {          
            \includegraphics[width=0.48\textwidth]{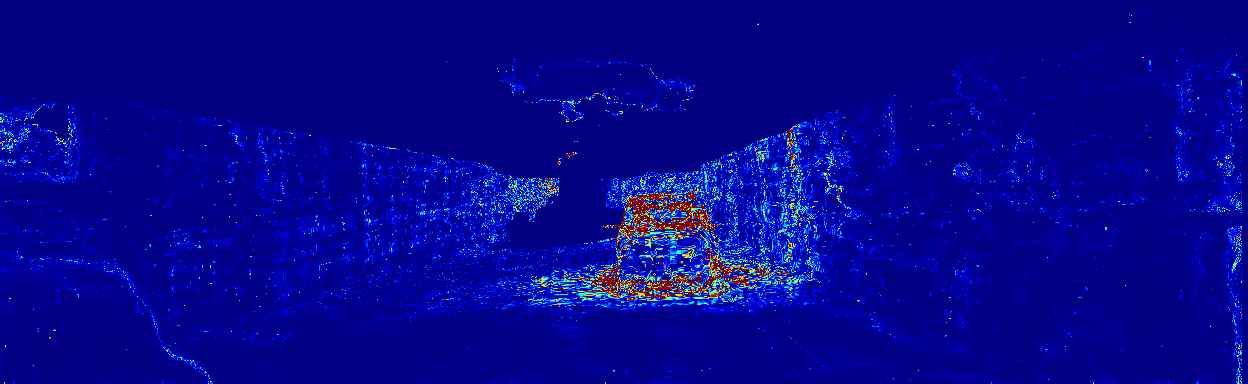}
         }    
         \\ 
         \subfigure
         {           
            \includegraphics[width=0.48\textwidth]{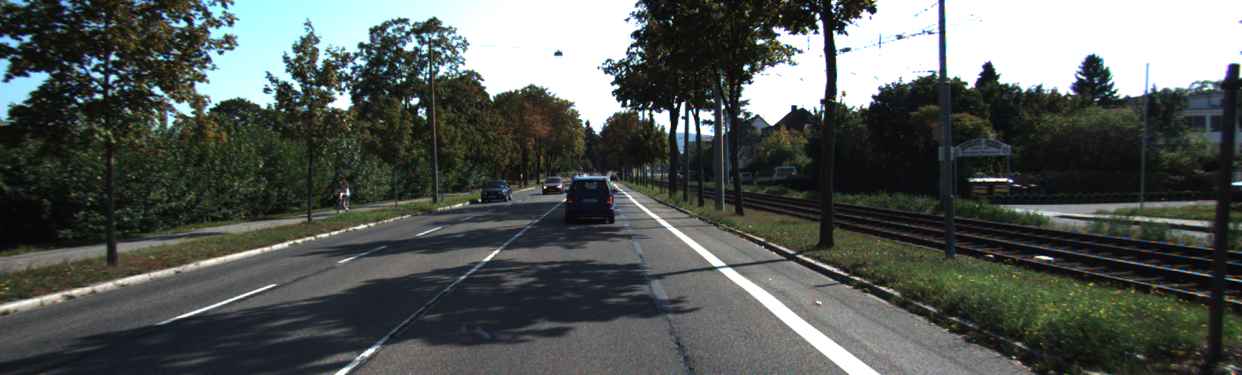}
         }
         \subfigure
         {          
            \includegraphics[width=0.48\textwidth]{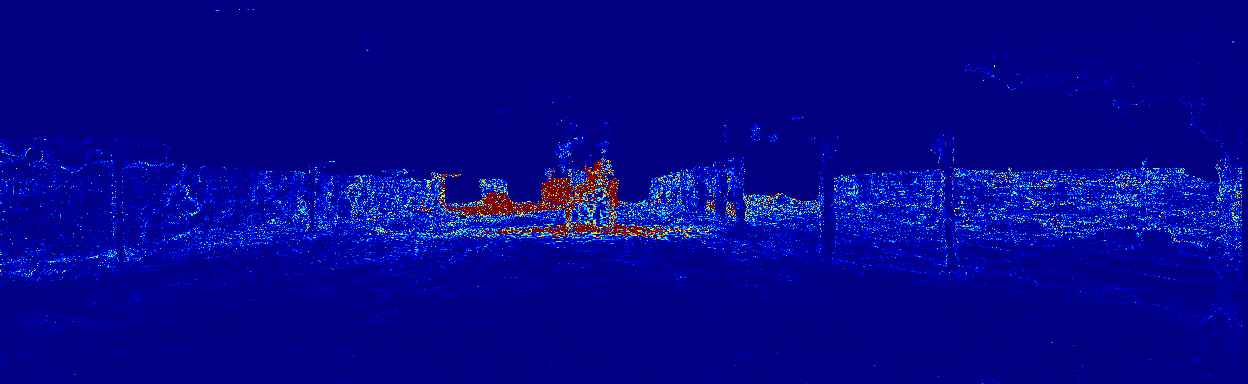}
         }\addtocounter{subfigure}{-4}
         \\ 
         \subfigure[Image]
         {           
            \includegraphics[width=0.48\textwidth]{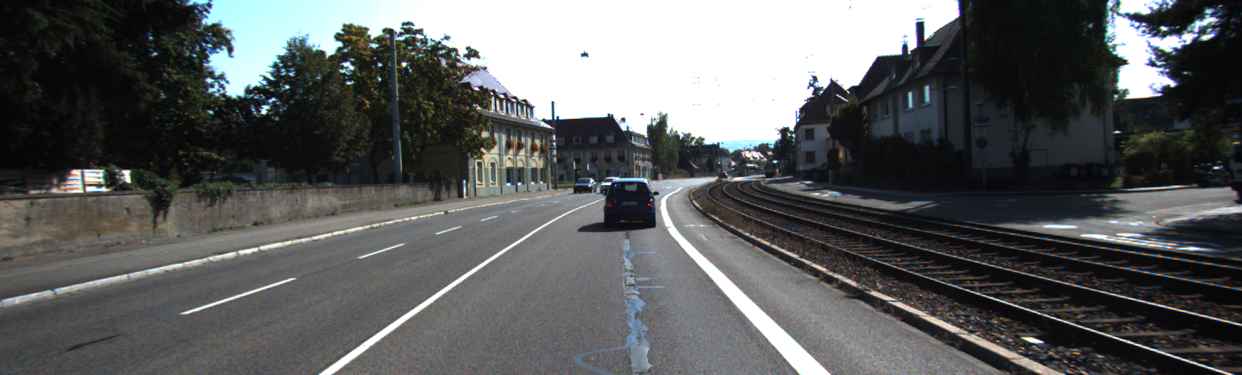}
         }
         \subfigure[Gradient from detector]
         {          
            \includegraphics[width=0.48\textwidth]{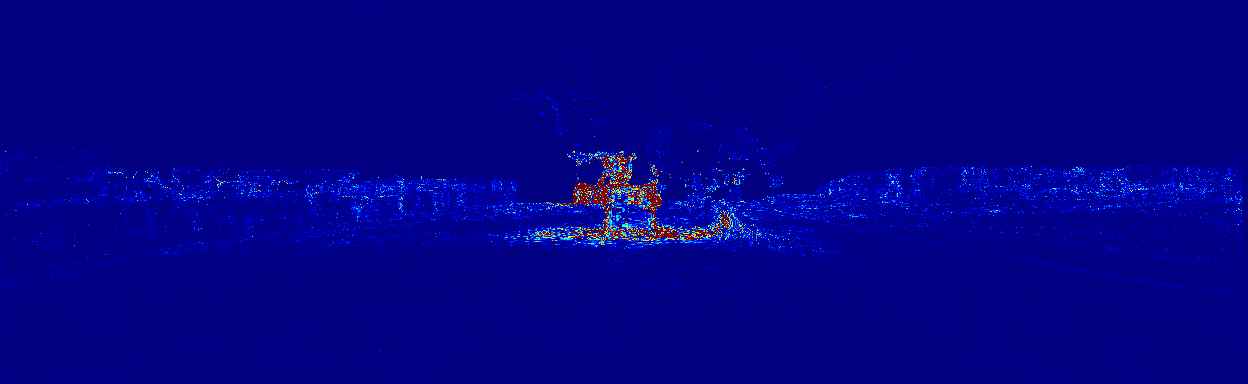}
         }    
      } 
   \end{center}
   \caption{\textbf{Visualization of absolute gradient values by the detection loss.} We use JET colormap to indicate the relative absolute values of resulting gradients, where red color indicates larger values; blue, otherwise.}
   \label{fig:cmp4}
\end{figure*}

We also visualize the gradients of the detection loss with respect to the depth map to indicate the effectiveness of our E2E-PL pipeline, as illustrated in \autoref{fig:cmp4}. We use JET Colormap to indicate the relative absolute value of gradients, where red color indicates higher values while blue color indicates lower values. The gradients from the detector focus heavily around cars.

\section{Other results}
\label{suppl-sec:other}

\subsection{Depth estimation}
We summarize the quantitative results of depth estimation (w/o or w/ end-to-end training) in \autoref{depth}. As the detection loss only provides semantic information to the foreground objects, which occupy merely $10\%$ of pixels (\autoref{fg:car_occ}), its improvement to the overall depth estimation is limited. But for pixels around the objects, we do see improvement at certain depth ranges. We hypothesize that the detection loss may not directly improve the metric depth, but will sharpen the object boundaries in 3D to facilitate object detection and localization.

\begin{table}[hbtp]
\centering
\begin{tabular}{c|c|c|c}
Range(meters)          & Model  & Mean error & Std\\ \hline
\multirow{2}{*}{0-10} & PL++    & 0.728  &  2.485 \\
                       & E2E-PL & \textbf{0.728} &  \textbf{2.435}  \\ \hline
\multirow{2}{*}{10-20} & PL++    & 1.000  &  3.113 \\
                       & E2E-PL & \textbf{0.984} &  \textbf{2.926}   \\ \hline
\multirow{2}{*}{20-30} & PL++    & 2.318  &   4.885\\
                       & E2E-PL & \textbf{2.259}  & \textbf{4.679} \\ \hline
\end{tabular}
\caption{Quantitative results on depth estimation.}
\label{depth}
\end{table}

\begin{table*}
\centering
\begin{tabular}{l|c|c|c|c|c|c|c}
\multicolumn{1}{c|}{\multirow{2}{*}{Method}} & \multirow{2}{*}{Input} & \multicolumn{3}{c|}{IoU=0.5}            & \multicolumn{3}{c}{IoU=0.7}                                        \\ \cline{3-8} 
\multicolumn{1}{c|}{}                        &                        & Easy        & Moderate    & Hard        & Easy                                    & Moderate    & Hard        \\ \hline
PL++: \PRCNN                  & S                      & 68.7 / 55.6 & 46.3 / 36.6 & 43.5 / 35.1 & 17.2 / \phantom{0}6.9 & 17.0 / 11.1 & 17.0 / 11.6 \\
\color{blue}
\ETE: \PRCNN   & \color{blue} S                      & \color{blue}\textbf{73.6 / 61.3} & \color{blue}\textbf{47.9 / 39.1} & \color{blue}\textbf{44.6 / 35.7} & \color{blue}\textbf{30.2 / 16.1}                             & \color{blue}\textbf{18.8 / 11.3} & \color{blue}\textbf{17.9} / 11.5 \\ \hline
\color{gray}
\PRCNN                        & \color{gray}L                      & \color{gray}93.2 / 89.7 & \color{gray}85.1 / 79.4 & \color{gray}84.5 / 76.8 & \color{gray}73.8 / 42.3                             & \color{gray}66.5 / 34.6 & \color{gray}63.7 / 37.4 \\ \hline
\end{tabular}
\caption{\textbf{3D object detection via the point-cloud-based pipeline with \PRCNN on Argoverse dataset.} We report \APBEV ~/ \AP (in \%) of the \textbf{car} category, using \PRCNN for detection. We arrange methods according to the input signals: S for stereo images, L for 64-beam LiDAR. PL stands for \PL. \emph{Results of our end-to-end \PL are in {\color{blue} blue}.} Methods with 64-beam LiDAR are in {\color{gray} gray}. Best viewed in color. \label{tb::argo}}
\end{table*}
\subsection{Argoverse dataset~\cite{argoverse}}
We also experiment with Argoverse~\cite{argoverse}. We convert the Argoverse dataset into KITTI format, following the original split, which results in $6,572$ and $2,511$ scenes (\ie, stereo images with the corresponding synchronized LiDAR point clouds) for training and validation. We use the same training scheme and hyperparameters as those in KITTI experiments, and report the validation results in \autoref{tb::argo}. We define the easy, moderate, and hard settings following \cite{yan2020train}. Note that since the synchronization rate of stereo images in Argoverse is 5Hz instead of 10Hz, the dataset used here is smaller than that used in \cite{yan2020train}. Our \ETE pipeline outperforms PL++. We note that, most existing image-based detectors only report results on KITTI.